\pdfoutput=1
\documentclass[lettersize,journal]{IEEEtran}
\usepackage{amsmath,amsfonts}
\usepackage{algorithmic}
\usepackage{algorithm}
\usepackage{array}
\usepackage[caption=false,font=normalsize,labelfont=sf,textfont=sf]{subfig}
\usepackage{textcomp}
\usepackage{stfloats}
\usepackage{url}
\usepackage{verbatim}
\usepackage{graphicx}
\usepackage{cite}
\usepackage{colortbl}
\begin{document}

\title{VQ-NeRF: Neural Reflectance Decomposition and Editing with Vector Quantization}

\author{Hongliang Zhong, Jingbo Zhang, and Jing Liao*
\thanks{*: corresponding author.}
\thanks{H. Zhong, J. Zhang and J. Liao are with Department of Computer Science, City University of Hong Kong. Email: hlzhong2-c@my.cityu.edu.hk, jbzhang6-c@my.cityu.edu.hk, jingliao@cityu.edu.hk.}
}

\maketitle

\begin{abstract}

We propose VQ-NeRF, a two-branch neural network model that incorporates Vector Quantization (VQ) to decompose and edit reflectance fields in 3D scenes. Conventional neural reflectance fields use only continuous representations to model 3D scenes, despite the fact that objects are typically composed of discrete materials in reality. This lack of discretization can result in noisy material decomposition and complicated material editing. To address these limitations, our model consists of a continuous branch and a discrete branch. The continuous branch follows the conventional pipeline to predict decomposed materials, while the discrete branch uses the VQ mechanism to quantize continuous materials into individual ones. By discretizing the materials, our model can reduce noise in the decomposition process and generate a segmentation map of discrete materials. Specific materials can be easily selected for further editing by clicking on the corresponding area of the segmentation outcomes. Additionally, we propose a dropout-based VQ codeword ranking strategy to predict the number of materials in a scene, which reduces redundancy in the material segmentation process. To improve usability, we also develop an interactive interface to further assist material editing. We evaluate our model on both computer-generated and real-world scenes, demonstrating its superior performance. To the best of our knowledge, our model is the first to enable discrete material editing in 3D scenes.

\end{abstract}

\section{Introduction}

Decomposing a scene into its constituent geometry, material, and lighting properties holds immense significance across various applications in the fields of computer vision and graphics, including scene relighting and appearance editing \cite{zhang2022modeling,kang2021neural,zhang2021nerfactor,boss2021nerd,wang2023neural}. This challenging task, commonly referred to as inverse rendering \cite{kim2019planar,yang2022ps,meka2018lime}, is inherently ill-posed due to the complex interplay between an object's observed color and its underlying lighting, material, and geometry attributes. For instance, the appearance of blackness in object renderings may be attributed to either insufficient lighting or dark material color. To overcome this inherent ambiguity, traditional methods for inverse rendering often incorporate additional constraints during the process of reflectance decomposition, such as controlled illumination \cite{sato1997object, dana2001brdf} or visual priors learned from 2D images \cite{li2017modeling, li2018learning, boss2020two}. However, such constraints significantly limit the applicability of the methods in real-world scenarios, where materials and illuminations are diverse and uncontrollable. In contrast, a more practical and universal strategy for inverse rendering is to introduce multi-view constraints and model the scene in a view-consistent representation \cite{joy2022multi, li2023multi}. This approach allows the model to analyze the appearance of the scene from multiple viewpoints, which can help to disambiguate the influence of lighting, material, and geometry, resulting in improved accuracy of reflectance decomposition. 

\begin{figure}
\centering
\includegraphics[width=0.99\linewidth]{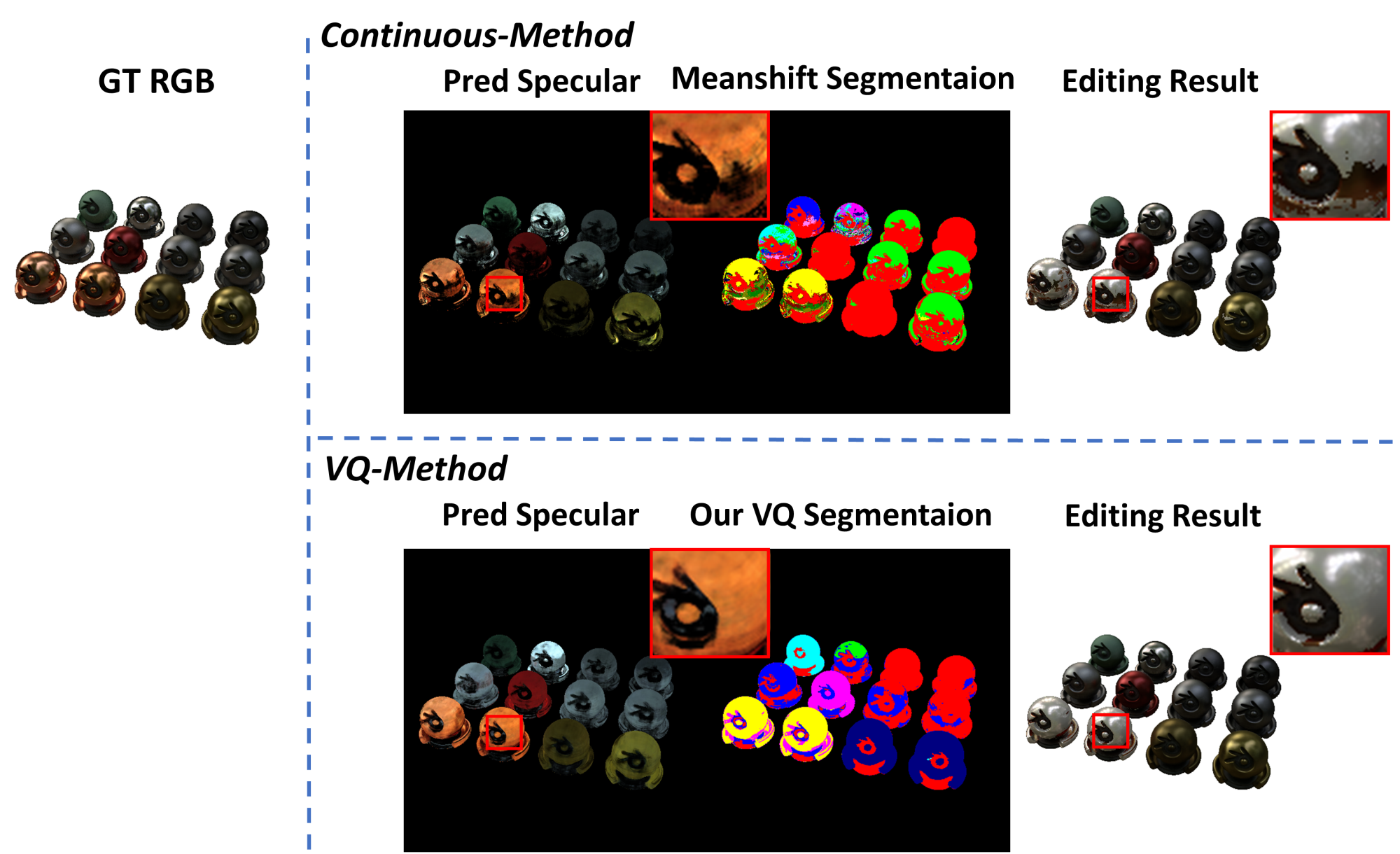}
\caption{Conventional neural reflectance decomposition methods (upper row) often predict noisy BRDF attributes for individual materials due to the absence of material discretization. This continuous representation also presents challenges for specific material editing. In contrast, our VQ-NeRF approach (lower row) incorporates the VQ mechanism to discretize reflectance decomposition, which suppresses prediction noise and facilitates material editing.}
\label{fig1}
\end{figure}

With the development of Neural Radiance Field (NeRF) \cite{mildenhall2021nerf}, recent reflectance decomposition methods (e.g., Bi et al. \cite{bi2020neural}, Zhang et al. \cite{zhang2021nerfactor}, Boss et al. \cite{boss2021nerd}, and Srinivasan et al. \cite{srinivasan2021nerv}) begin to adopt it as a 3D representation and introduce multi-view constraints. Briefly, these neural reflectance decomposition methods learn a continuous field that maps the spatial coordinates of the scene to corresponding reflectance factors represented by a Spatially-Varying Bidirectional Reflectance Distribution Function (SV-BRDF) \cite{guo2018brdf}. Thanks to the powerful modeling and rendering capabilities of the neural implicit field, these methods can capture the subtle characteristics in the texture and geometry from different views of the scene, resulting in more accurate decomposition results in the inverse rendering task. However, their continuous representation of BRDF attributes conflicts with reality, where objects are typically composed of discrete types of materials such as wood, plastic, metal, and others. The absence of discretization often leads to noisy decomposition within individual materials. As shown in the upper row of Fig. \ref{fig1}, the predicted specular attributes for the \textit{bronze} balls vary significantly from location to location. Furthermore, the non-discrete modeling makes selecting specific materials for editing challenging, which in turn complicates appearance editing. Even with the help of the Meanshift clustering, as illustrated in Fig. \ref{fig1}, selecting the entire ball with the \textit{bronze} material and editing it into a new \textit{silver} material remains a challenge for such continuously decomposed materials.

To address the aforementioned issue, we propose VQ-NeRF, a novel neural reflectance decomposition framework based on Vector Quantization (VQ) \cite{van2017neural, yu2021vector}. Our framework comprises a continuous branch and a discrete branch. The continuous branch predicts a 3D implicit field of reflectance factors under multi-view constraints, while the discrete branch employs the VQ mechanism to quantize the continuous reflectance field into a limited number of VQ codewords, resulting in a discrete segmentation map for different materials. These two branches are jointly optimized, with the VQ clustering in the discrete branch effectively constraining the reflectance prediction of the continuous branch to be more compact, approaching VQ codewords, and thus suppressing prediction noise, as illustrated in the lower row of Fig. \ref{fig1}. In turn, the adjusted prediction results in the continuous branch assist the discrete branch in learning more accurate codewords, enhancing material clustering performance.

\begin{figure*}
\centering
\includegraphics[width=0.99\linewidth]{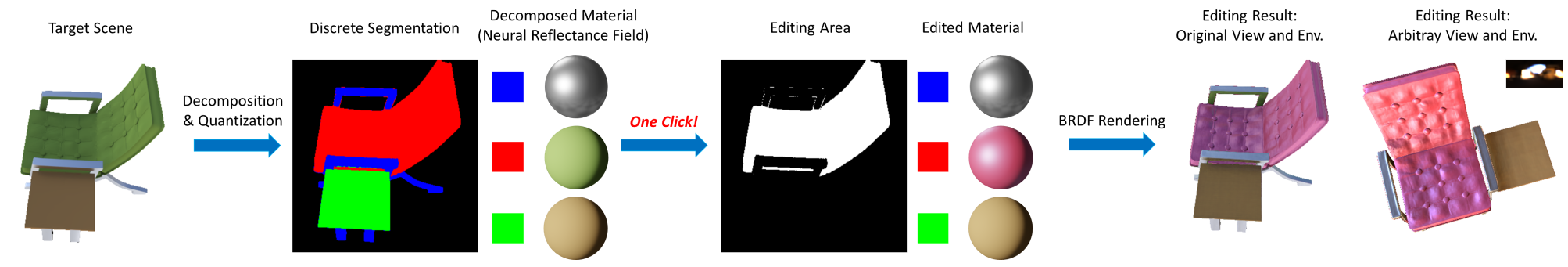}
\caption{We propose VQ-NeRF, which incorporates the VQ mechanism to discretize reflectance decomposition. This enables efficient and view-consistent material selection and editing.}
\label{fig0}
\end{figure*}

Moreover, our VQ-NeRF framework enables the user to conveniently select and edit specific materials by providing a material segmentation map through the discrete branch, as shown in Fig. \ref{fig0}. To prevent the presence of redundant materials in the segmentation map, we further introduce a dropout-based codeword ranking strategy to our VQ scheme. By sorting the material codewords according to their importance, our method can eliminate lower-ranked redundant codewords, ensuring that the number of predicted materials is appropriate for the scene complexity and facilitating user selection and editing of specific materials. Additionally, to support intuitive material editing in 3D scenes, we have built an interactive User Interface (UI). In this interface, users can view the segmentation map produced by the discrete branch from any angle and click on the corresponding areas to specify the materials to be edited. Our continuous branch then performs neural rendering with the edited materials and presents the results in different views. This allows users to have a more intuitive and interactive experience when editing materials in 3D scenes.

Thanks to the well-designed two-branch framework with VQ mechanism and dropout-based ranking strategy, our VQ-NeRF significantly enhances the accuracy of reflectance decomposition and empowers efficient material editing. We evaluated our method on both Computer-Generated (CG) scenes and real-world scenes, and demonstrated its superior performance in multiple tasks including scene reconstruction, reflectance decomposition, material editing, and scene relighting.

To sum up, our contributions are three folds:
\begin{itemize}
\item We propose VQ-NeRF, the first method that incorporates the VQ mechanism to discretize reflectance decomposition, thereby enhancing decomposition accuracy and facilitating material editing.
\item We introduce a dropout-based VQ codeword ranking strategy that automatically determines the number of materials in an arbitrary scene, eliminating redundancy in VQ-predicted materials.
\item We develop an interactive user interface that enables convenient material editing of 3D scenes in a view consistent manner.
\end{itemize}

\section{Related Work}
\subsection{Traditional Reflectance Decomposition}
Considering the inherent ambiguity during inverse rendering, classical methods usually require additional constraints to assist the reflectance decomposition process of the model \cite{wu2015simultaneous, deschaintre2018single, deschaintre2019flexible, boss2020two}. For example, LSR-BRDF \cite{sato1997object} and CM-BRDF \cite{dana2001brdf} simplify the reflectance computation by capturing scenes in a controllable lighting environment. Although they achieve plausible decomposed results in their specific experimental scenes, they cannot perform reasonable reflectance estimation for realistic scenarios with arbitrary and uncontrollable illumination. In contrast to constrain the lighting conditions, SA-CNN \cite{li2017modeling} adopts a deep convolution network to learn visual priors from planar material data and performs SV-BRDF estimation in the image space. Benefiting from the data-driven priors, this method achieves 2D reflectance decomposition under casual illumination. Nonetheless, it struggles to handle reflectance decomposition of 3D objects due to the lack of 3D priors. By contrast, CASCADE-CNN \cite{li2018learning} trains a cascaded network on a more general 3D dataset containing multi-view material factors rendered from 3D objects using a complex SV-BRDF. Thanks to the enhanced model and 3D priors in the training data, this method enable to infer reflectance factors from rendered images of 3D objects. Still, it fails to produce view-consistent decomposed factors since the inference of each view is performed independently. To introduce multi-view constraints, MVGCR \cite{joy2022multi} reconstructs polygon meshes of the scene with Multi-View Stereo (MVS) \cite{schonberger2016structure}, and uses it as a 3D representation to perform inverse rendering. Although such method realizes view-consistent reflectance decomposition, the low-fidelity mesh representation seriously limits its performance.

\subsection{Neural Reflectance Decomposition}
\label{cmp_models}

Inspired by the great success of the emerging NeRF and its variants \cite{mildenhall2021nerf, chen2022tensorf, barron2021mip} in 3D scene modeling, recent methods of reflectance decomposition \cite{srinivasan2021nerv, boss2021nerd, zhang2021physg, kuang2022neroic} attempt to leverage neural implicit fields \cite{wang2022nerfcap, chen2022tensorf} as 3D representations to provide multi-view constraints during inverse rendering. For instance, NeRFactor \cite{zhang2021nerfactor} utilizes multiple implicit fields to model the scene geometry, albedo, and BRDF identity, respectively. Benefiting from the powerful representation capability of neural implicit fields in view-consistent modeling of materials, NeRFactor demonstrates promising results compared to traditional methods. However, since NeRFactor predicts specular reflections through a pre-trained network, its decomposition is largely constrained by the distribution bias of the pre-training data. Similarly, Neural-PIL \cite{boss2021neural} adopts a pre-trained network for material prediction and a pre-trained network for lighting estimation. This design allows Neural-PIL to predict different lighting in different views and thus achieve reflectance decomposition for scenes rendered in varying illuminations. Nonetheless, it suffers the same limitation as NeRFactor, as the bias of pre-training data seriously limits its decomposition performance. By contrast, NeILF \cite{yao2022neilf} introduces an additional implicit lighting field to predict a corresponding lighting intensity for each surface point of the object. In this way, it is theoretically able to model the indirect lighting, leading to enhanced processing of complex illumination. However, due to the lack of sufficient restrictions, during the actual material decomposition process, this lighting field may be confused with the material fields, causing the lighting colors to be absorbed into the material fields. Unlike previous methods that employ volume rendering during optimization, NVDIFFREC \cite{munkberg2022extracting} and NVDIFFRECMC \cite{hasselgren2022shape} incorporate traditional rasterization-based rendering into their framework to accelerate computation by extracting scene meshes from their signed distance fields. However, explicit rasterization can easily lead to visible artifacts in rendered images, such as stretched geometry and blurry textures, which in turn affects the decomposition quality of materials, especially for some geometrically complex scenes. Besides, to best of our knowledge, all of existing methods for neural reflectance decomposition focus on continuous BRDF estimation, which is not conducive to material selection and editing, and conflicts with reality, because objects in real scenes are usually composed of discrete types of material. Therefore, we design a two-branch neural reflectance field based on the VQ mechanism to achieve discrete BRDF material decomposition.

\begin{figure*}
\centering
\includegraphics[width=0.99\linewidth]{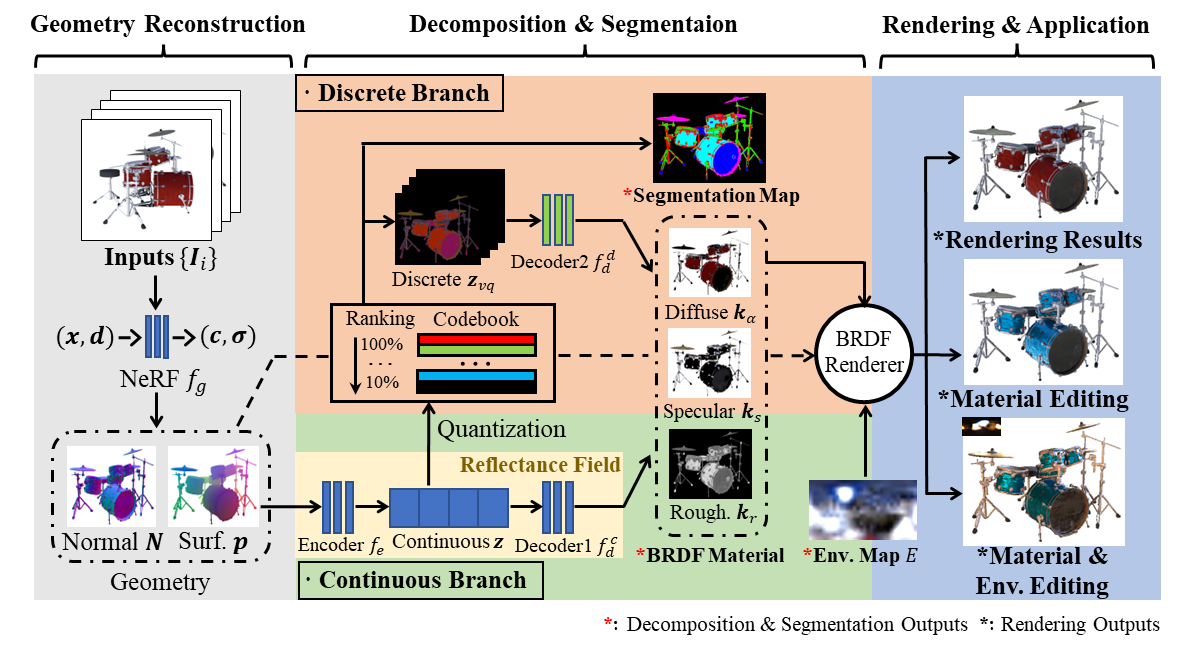}
\caption{The pipeline of our VQ-NeRF, the outputs are marked by asterisks (*). We first take multi-view posed images as inputs and use a NeRF model (gray part) to reconstruct the scene geometry. Next, we apply a two-branch network for reflectance decomposition and material discretization. The continuous branch (green part) predicts the decomposed BRDF attributes, including diffuse, specular, and roughness, while the discrete branch (red part) uses the VQ mechanism to discretize reflectance factors. After optimization, a material segmentation map is generated, which enables us to easily select specific materials for editing.}
\label{fig2}
\end{figure*}

\section{Method}
\label{sec_method}
The pipeline of our VQ-NeRF approach is illustrated in Fig. \ref{fig2}. We first use a NeRF model to reconstruct the scene geometry and extract geometry components, such as surface normals and coordinates. Next, we jointly optimize a continuous branch network and a discrete branch network to perform reflectance decomposition and VQ-based material discretization, respectively. The continuous branch learns a neural reflectance field that maps spatial coordinates of the scene to corresponding reflectance factors represented by SV-BRDF, including diffuse, specular, and roughness, as well as an environment map. Meanwhile, the discrete branch employs the VQ mechanism to quantize the continuous reflectance field into a limited number of VQ codewords, yielding a material segmentation map. Additionally, we apply a dropout-based codeword ranking strategy to the discrete branch to reduce quantization redundancy. With the view-consistent segmentation map in an arbitrary rendering view, users can easily select a specific material for editing and produce the edited scene using BRDF rendering.

\subsection{Geometry Reconstruction}

The inputs to our method is a set of posed images ${I_{i}}$, which are captured from a 3D scene under natural illumination. To perform inverse rendering, we first use a NeRF network $f_{g}$ to reconstruct the scene geometry. 
Similar as previous work, volume rendering \cite{mildenhall2021nerf} is employed to accumulate the color $C_{v}(\boldsymbol{r})$ in NeRF:
\begin{equation}
    \begin{cases}
      C_{v}(\boldsymbol{r}) =\int_{t_{n}}^{t_{f}}T(t)\sigma(\boldsymbol{r}(t))
      c(\boldsymbol{r}(t), \boldsymbol{d}){dt}, \\
      T(t)=\exp(-\int_{t_{n}}^{t}\sigma(\boldsymbol{r}(s)){ds}),
    \end{cases}
\end{equation}
where $\boldsymbol{r}(t)=\boldsymbol{o}+t\boldsymbol{d}$ is the spatial points sampled on the camera ray emitted from the origin $\boldsymbol{o}$ in the direction $\boldsymbol{d}$. $t_n$ and $t_f$ represent the bounds of near and far sampling. $\sigma(\boldsymbol{r}(t))$ and $c(\boldsymbol{r}(t), \boldsymbol{d})$ indicate the predicted density and color of the sampled point $\boldsymbol{r}(t)$, respectively.

To train this network, we minimize the $L_2$ loss between the rendered color $C_{v}(\boldsymbol{r})$ and the pixel color $C_{gt}(\boldsymbol{r})$ in the input images. After training, we can extract the geometry components from the reconstructed model $f_{g}$, including the coordinates $\boldsymbol{p}$ of surface points and the associated surface normal $\boldsymbol{N}(\boldsymbol{p})$ \cite{zhang2021nerfactor,boss2021nerd}.

\subsection{Continuous Branch}
\label{sec_continue}

We construct a neural reflectance field in the continuous branch to map surface coordinates $\boldsymbol{p}$ into BRDF attributes. The reflectance field involves an encoder $f_e$ and a decoder $f_d^c$, both of which are composed of MLP networks. The encoder maps the input spatial coordinates $\boldsymbol{p}$ into latent material vectors $\boldsymbol{z}(\boldsymbol{p})$ and the decoder predicts the BRDF material factors according to the latent vectors $\boldsymbol{z}(\boldsymbol{p})$:
\begin{equation}
\begin{cases}
        \boldsymbol{k}_d(\boldsymbol{p}), \boldsymbol{k}_m(\boldsymbol{p}), \boldsymbol{k}_r(\boldsymbol{p}) = f_d^c(\boldsymbol{z}(\boldsymbol{p})), \\
        \boldsymbol{z}(\boldsymbol{p})=f_e(\boldsymbol{p}),
\end{cases}
\end{equation}
where $\boldsymbol{k}_d$, $\boldsymbol{k}_m$, and $\boldsymbol{k}_r$ indicate the basecolor, metallic, and roughness, respectively \cite{yao2022neilf, munkberg2022extracting}. Subsequently, the basecolor $\boldsymbol{k}_d$ and metallic $\boldsymbol{k}_m$ are further converted into the diffuse attribute $\boldsymbol{k}_\alpha$ and the specular attribute $\boldsymbol{k}_s$ for the following rendering \cite{boss2021neural}:
\begin{equation}
\begin{cases}
    \boldsymbol{k}_\alpha(\boldsymbol{p}) = \boldsymbol{k}_d(\boldsymbol{p}) \cdot (1 - \boldsymbol{k}_m(\boldsymbol{p})), \\
    \boldsymbol{k}_s(\boldsymbol{p}) = \boldsymbol{k}_d(\boldsymbol{p}) \cdot \boldsymbol{k}_m(\boldsymbol{p}).
\end{cases}
\end{equation}

To deduce the rendered color $C_{r}(\boldsymbol{p})$ of the surface point $\boldsymbol{p}$ from these predicted BRDF attributes, we adopt Microfacet BRDF model \cite{kajiya1986rendering} as our BRDF renderer:
\begin{equation}
    C_{r}(\boldsymbol{p}) = \int_{\Omega}{
    L_{i} \cdot
    f_{R}(\boldsymbol{k}_\alpha, \boldsymbol{k}_s, 
    \boldsymbol{k}_r;
    \boldsymbol{p}, \boldsymbol{\omega}_i, 
    \boldsymbol{\omega}_o)
    (\boldsymbol{\omega}_i 
    \boldsymbol{N}(\boldsymbol{p}))
    }d{\boldsymbol{\omega}_i},
  \label{eq:rendering}
\end{equation}
where $\Omega$ represents the illumination sphere surrounding the scene. $L_{i}$ denotes the incoming illumination from the $i$-th lighting source, which is sampled from a learnable environment map $E$. $f_R(\cdot)$ denotes the BRDF function. $\boldsymbol{\omega}_{i}$ and $\boldsymbol{\omega}_{o}$ represent the lighting direction and viewing direction, respectively. $(\boldsymbol{\omega}_i \boldsymbol{N}(\boldsymbol{p}))$ indicates the angle between the surface normal $\boldsymbol{N}(\boldsymbol{p})$ and the lighting direction $\boldsymbol{\omega}_{i}$. 

Due to the BRDF attributes $\boldsymbol{k}_d$, $\boldsymbol{k}_m$, and $\boldsymbol{k}_r$ are predicted from the continuous neural reflectance field, it conflicts with reality where the material distribution in the scene is discretized and regionalized. 
To solve this issue, we introduce a discrete branch in the following section to quantize the hidden vectors predicted by the continuous branch and produce discretized material attributes.

\subsection{Discrete Branch}
\label{sec_disc}

\subsubsection{Vector Quantization}
\label{vq-sec}
In this section, we construct a discrete branch combined the VQ mechanism \cite{van2017neural, yu2021vector} for material discretization, which facilitates material selection and editing. For each latent material vector $\boldsymbol{z}$ produced by the continuous branch, VQ mechanism matches it with a most similar codeword $\boldsymbol{z}_{vq}$ selected from its trainable codebook:

\begin{equation}
    \begin{cases}
      u = {\underset{i}{\text{argmin}} \, |\boldsymbol{e}_i - \boldsymbol{z}|^{2}}, ~ i \in {1,...,M}, \\
      \boldsymbol{z}_{vq} = \text{sg}(\boldsymbol{e}_u - \boldsymbol{z}) + \boldsymbol{z},
      \end{cases}
  \label{eq:vq}
\end{equation}
where $M$ denotes the length of the VQ codebook. $\boldsymbol{e}_i$ repesents the $i$-th codeword. $\text{sg}(\cdot)$ indicates the stop gradient operation. During the VQ clustering, both $\boldsymbol{z}$ and $\boldsymbol{z}_{vq}$ are normalized to the unit sphere for computational convenience. 

Consequently, we employ another decoder $f_d^d$ to infer discrete material attributes from the quantized latent vectors $\boldsymbol{z}_{vq}$. With these discrete material attributes, a similar BRDF rendering process can be performed to infer the rendered color $C_{r,d}(\boldsymbol{p})$ in arbitrary views by using Eq. \ref{eq:rendering}. Besides, we can easily deduce a material segmentation map according to these discrete materials for facilitating material selection and editing. Notably, since both $\boldsymbol{z}$ and $\boldsymbol{z}_{vq}$ are predicted solely from the surface points $\boldsymbol{p}$, the material segmentation map is view-consistent and can be deduced in any desired view.

\begin{figure}
\centering
\includegraphics[width=0.95\linewidth]{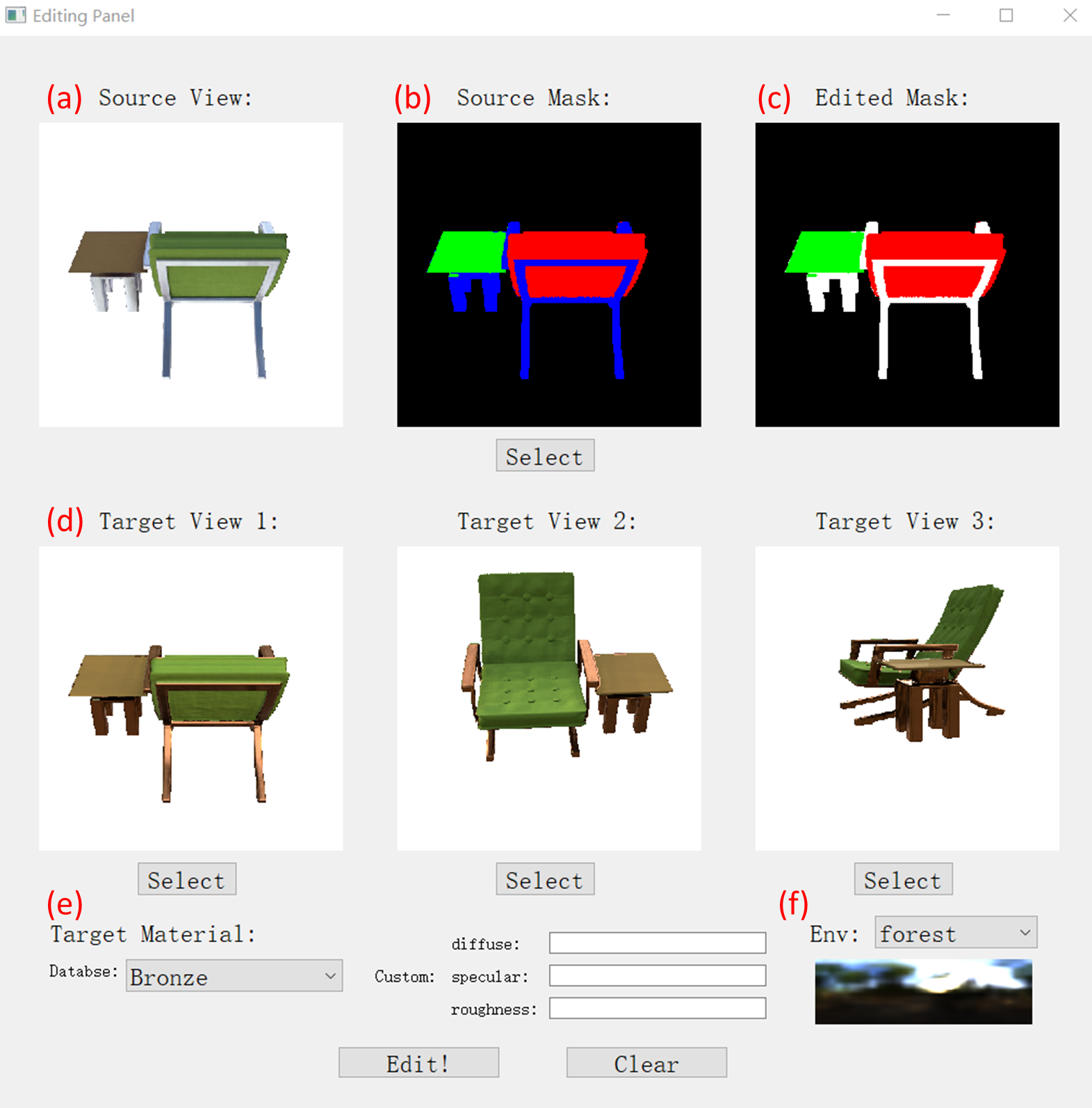}
\caption{Our UI for interactive material selection and editing. The reconstructed image and segmentation map of an arbitrary view are presented in (a), (b) and (c). Users can click in (b) to specify the editing area, and assign the target material in (e). The lighting of the scene can also be adjusted in (f). After configuring all the settings, the user can start the model by clicking the 'Edit!' button. The edited results are shown in (d) for visualization.}
\label{fig16}
\end{figure}

\subsubsection{Dropout-based Ranking}
Although VQ achieves the discretization of continuous materials, how to determine the length of the VQ codebook $M$ is still a problem due to the uncertainty of the number of materials in a scene. To determine the codebook length automatically and eliminate material redundancy, we introduce a dropout-based codeword ranking strategy during the reflectance decomposition. Specifically, we first set an initial length $M_0$ for the codebook and assign a dropout rate to each codeword. The dropout rates are set in ascending order, increasing linearly from $0$ to around $0.7$. During optimization, the codewords will be randomly dropped out from the codebook according to their assigned dropout rates. The frontier codewords, which possess lower dropout rates, are more likely to participate in model optimization and loss computation. As a result, to minimize the overall reconstruction loss of the discrete branch, the frontier codewords are considered crucial materials that significantly impact the reduction of the reconstruction error. Through this importance-driven prediction, the codewords are automatically sorted by importance after training.

Then, we perform multiple evaluations to determine the appropriate length $M$ for the codebook. Specifically, we conduct the decomposition and BRDF rendering process in the discrete branch multiple times, using the first $k$ codewords from the codebook, and compute the reconstruction error $err_{k}$ at each evaluation. As the codewords have been sorted by importance, the curve formed by these reconstruction errors exhibits a trend of rapid decline followed by flattening. Therefore, we can determine the length $M$ by finding the point on the curve where the flattening occurs, i.e., the first point satisfying the condition $ \left|err_k - err_{i}\right|<=\epsilon, i \in {k+1,...,M_0}$, from $k=1$ to $M_0$. Here, $\epsilon$ is a constant for flattening determination, which is empirically set as $0.002$ in practice.

\subsection{Two-branch Joint Training}
\label{train-sec}
To encourage mutual benefit between the continuous and discrete branches, we use a joint training strategy. During training, we adopt compound objectives to constrain our two-branch reflectance decomposition framework, including a two-branch reconstruction loss, a VQ loss, a smooth loss, and a Lambertian loss. 

Specifically, we separately calculate the $L_2$ losses between the rendered color $C_{r}$ ($C_{r,d}$) and the pixel color $C_{gt}$ in the continuous and discrete branches as the reconstruction loss. Besides, to eliminate the influence of illumination on the VQ discretization process, we also calculate the reconstruction loss of the discrete branch in the chromaticity space: $L_{chr} = |\text{chr}(C_{r,d}) - \text{chr}(C_{gt})|^{2}$, where $\text{chr}(\cdot)$ is the transformation function from RGB space to chromaticity space. Therefore, the two-branch reconstruction loss is defined as $L_{rec}=w_1 L_{rec,c}+w_2 L_{rec,d}+w_3 L_{chr}$. Here, $L_{rec,c}$ and $L_{rec,d}$ are the $L_2$ losses in the continuous and discrete branches, respectively. $w_1$, $w_2$, and $w_3$ are constant parameters balancing between terms. The VQ loss is composed of two terms, defined as:
\begin{equation}
  L_{vq} = |\boldsymbol{z}_{vq} - \text{sg}(\boldsymbol{z})|^{2} + \lambda\cdot{|\boldsymbol{z} - \text{sg}(\boldsymbol{z}_{vq})|^{2}},
  \label{eq:vqloss}
\end{equation}
where $\lambda$ is a constant parameter. Additionally, we design a Lambertian loss such that surface points with high roughness are predicted to have low specular, which is consistent with real material relations. The Lambertian loss is formed as $L_{lam} = w_{r} \cdot \boldsymbol{k}_s$, where the weight $w_{r} = 2 \cdot \text{sg}(\boldsymbol{k}_r) - 1$ if $\boldsymbol{k}_r>0.5$ else $0$. 

In practice, we find that a large area of the same material may be mistakenly divided into multiple pieces of similar materials during VQ clustering. To solve this problem, we introduce a smooth loss to achieve a color-aware constraint:
\begin{equation}
      L_{sm} = \text{exp}(-\alpha \cdot e_{chr}) \cdot (1 - \boldsymbol{z}_{vq,i} ,\cdot \boldsymbol{z}_{vq,j}),
  \label{eq:smoothloss}
\end{equation}
where the exponential component $e_{chr} = ||\text{chr}(C_{gt,i}) - \text{chr}(C_{gt,j})||_{2}^{2}$ if $||\text{chr}(C_{gt,i}) - \text{chr}(C_{gt,j})||_{2}^{2} > \beta$, else $e_{chr}=0$. $\alpha$ and $\beta$ are constant parameters for value scaling and threshold clipping. $C_{gt,i}$ and $C_{gt,j}$ indicate the colors of adjacent surface points $\boldsymbol{p}_i$ and $\boldsymbol{p}_j$.  $\boldsymbol{z}_{vq,i}$ and $\boldsymbol{z}_{vq,j}$ are the corresponding clustering codewords. 

Thus, the overall objective can be be expressed as $L_{all}=L_{rec}+w_{4}L_{vq}+w_{5}L_{lam}+w_{6}L_{sm}$, where $w_{4}$, $w_{5}$, and $w_{6}$ are constant parameters.

\subsection{User Interface}
Furthermore, we develop a UI for interactive material selection and editing. As shown in Fig. \ref{fig16}, users enable render the reconstructed model in an arbitrary view (a) and obtain the corresponding segmentation map (b) inferred by our discrete branch. Then, they can click the segmentation map to select the areas to be edited, as shown in (c). By setting the target material and environment map on (e) and (f) from the associated databases, the edited 3D model with desired material and illumination will be re-rendered in the region (d) of the UI.

\section{Experiments}
\label{exp}

\subsection{Setup}
\label{expset}
\noindent\textbf{Experiment Data.}
To evaluate the performance of our method, we conduct reflectance decomposition experiments on both CG dataset and real dataset. Here, we collect five scenes (\textit{drums}, \textit{hot-dog}, \textit{ficus}, \textit{lego}, and \textit{metal-balls}) released by NeRFactor \cite{zhang2021nerfactor} and NeRF \cite{mildenhall2021nerf} as the CG dataset. Notably, due to these scenes lack the ground truth of specular and roughness, we construct three additional scenarios (\textit{kitchen}, \textit{chair}, and \textit{blender}) for evaluation on specular and roughness components. The real dataset includes seven scenes captured by us (\textit{rabbit}, \textit{kettle}, \textit{tools}, \textit{shoes}, \textit{wooden-chair}, \textit{redcar}, and \textit{lord-rabbit}) and three scenes collected from the DTU \cite{wang2021neus} dataset (\textit{golden-sculpture}, \textit{house}, and \textit{dolls}).

\noindent\textbf{Baseline Methods.}
We compared our VQ-NeRF to five state-of-the-art reflectance decomposition methods, including NeRFactor \cite{zhang2021nerfactor}, Neural-PIL \cite{boss2021neural}, NeILF \cite{yao2022neilf}, NVDIFFREC \cite{munkberg2022extracting}, and NVDIFFRECMC \cite{hasselgren2022shape}. All of the experiments are conducted on author-released codes for compared methods.

\noindent\textbf{Metrics.}
Following other works \cite{mildenhall2021nerf,zhang2021nerfactor,munkberg2022extracting}, we use Peak Signal-to-Noise Ratio (PSNR), Structural Similarity Index Measure (SSIM), and Learned Perceptual Image Patch Similarity (LPIPS) as quantitative evaluation metrics in appearance reconstruction, reflectance decomposition, and scene relighting. Higher PSNR/SSIM scores and lower LPIPS scores indicate better quality. As for the evaluation of segmentation results, we follow \cite{lu2022mean} to measure segmentation accuracy using F1-score, Precision Rate (P) and Recall Rate (R) calculated under both micro and macro average. Higher scores for all of these metrics indicate better quality.

\noindent\textbf{Implementation Details.}
In practice, we set the super-parameters in training loss as 
$w_{1}=0.2$, $w_{2}=w_{3}=w_{4}=1$, $w_{5}=0.001$, and $w_{6}=0.05$. The $\lambda$ in Eq. \ref{eq:vqloss} is set to $0.1$. And for Eq. \ref{eq:smoothloss}, we set $\alpha=60$ and $\beta=0.1$. $M_0$ is set to $8$ for most of the scenes. But for the \textit{tools} scene, the \textit{kitchen} scene, and the scenes in the CG dataset, we set $M_0=15$, as their material compositions are more complicated. Our pipeline supports the use of various NeRF variants for geometry reconstruction. Specifically, we use NeuS \cite{wang2021neus} in our implementation. To achieve robust convergence, we employ VQ-EMA \cite{roy2018theory} instead of plain VQ in our discrete branch. Inspired by \cite{ye2022intrinsicnerf, wu2022palettenerf}, we additionally bake a residual into the reconstructed images. The residual baking is independent of reflectance decomposition, material editing, and scene relighting, but can bring richer details (such as intra-scene reflections) in appearance reconstruction. Considering the differences between the BRDF models adopted by different methods, the light-albedo scales of different methods are inconsistent. To solve this issue and perform a fair comparison, similar as previous methods \cite{zhang2021nerfactor, munkberg2022extracting, hasselgren2022shape}, we normalize the decomposed materials and relighted images to match the average luminance of the reference via an indeterminable scale factor for each method.

\begin{figure*}
\centering
\includegraphics[width=0.95\linewidth]{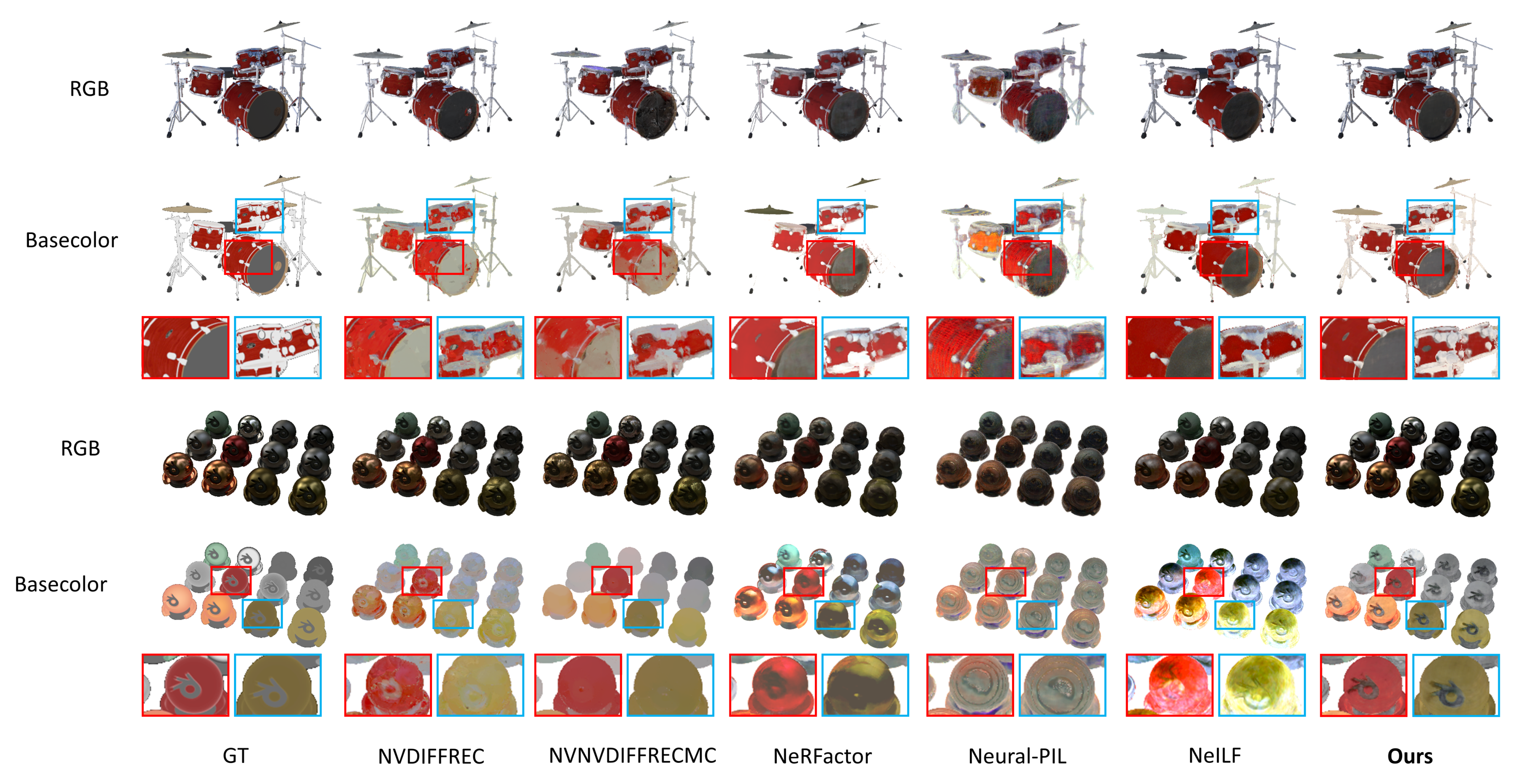}
\caption{Reconstruction and reflectance decomposition results on the CG dataset provided by previous methods. Obviously, the reflectance factors predicted by our model exhibit the most proper color and the least noise.}
\label{fig3}
\end{figure*}

\begin{table}
\caption{Reconstruction and reflectance decomposition results on the CG dataset provided by previous methods.}
\label{tab:nerfactor}
\centering
\tabcolsep=0.125cm
\begin{tabular}{c|ccc|ccc}
\hline
\multicolumn{1}{c|}{} & \multicolumn{3}{c|}{Reconstruction} & \multicolumn{3}{c}{Basecolor} \\
\hline
 & PSNR↑ & SSIM↑ & LPIPS↓ & PSNR↑ & SSIM↑ & LPIPS↓\\
NVDIFFREC & 33.193 & 0.968 & 0.021 & 25.006 & 0.926 & 0.080 \\
NVDIFFRECMC & 32.090 & 0.960 & 0.032 & 27.582 & {\bf 0.953} & 0.054 \\
NeRFactor & 31.736  & 0.947 & 0.037 & 26.452 & 0.942 & 0.057 \\
Neural-PIL & 26.366  & 0.899 & 0.084 & 24.298 & 0.890 & 0.112 \\
NeILF & 32.365 & 0.958 & 0.024 & 23.883 & 0.911 & 0.082 \\
{\bf Ours} & {\bf 35.390} & {\bf 0.975} & {\bf 0.014} & {\bf 29.475} & 0.949 & {\bf 0.046} \\
\hline
\end{tabular}
\end{table}

\begin{figure*}
\centering
\includegraphics[width=0.95\linewidth]{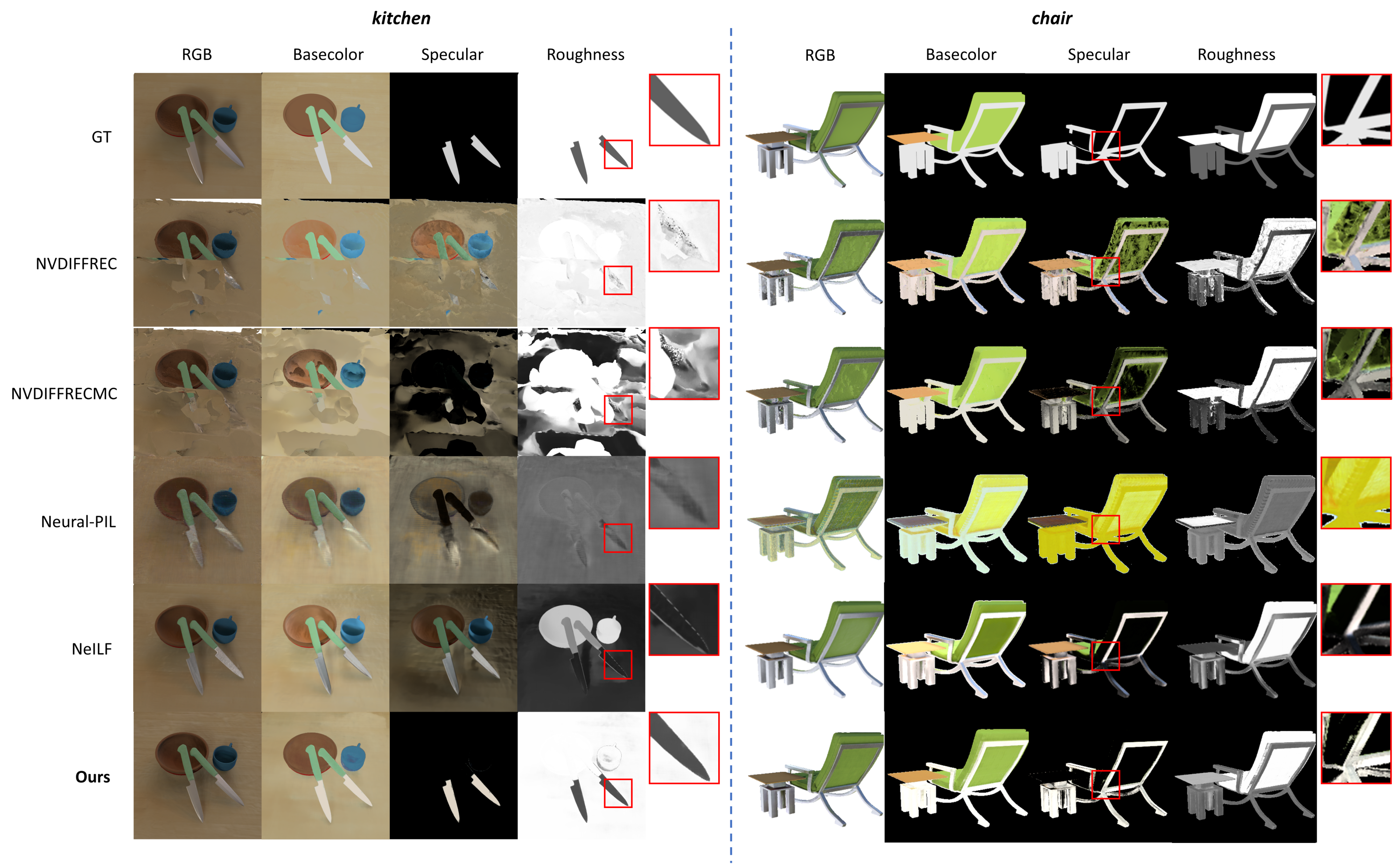}
\caption{Reconstruction and reflectance decomposition results on our CG dataset. The specular and roughness of NeRFactor are parameterized as latent codes in a network \cite{zhang2021nerfactor, zhang2023nemf} and cannot be extracted explicitly. So we exclude NeRFactor from this comparison. Evidently, our model achieves superior performance in the prediction of all three BRDF attributes.}

\label{fig4}
\end{figure*}

\begin{table*}
\caption{Reconstruction and reflectance decomposition results on our CG dataset}
\label{tab:material}
\centering
\begin{tabular}{c|ccc|ccc|ccc|ccc}
\hline
\multicolumn{1}{c|}{} & \multicolumn{3}{c|}{Reconstruction} & \multicolumn{3}{c|}{Basecolor} & \multicolumn{3}{c|}{Specular} & \multicolumn{3}{c}{Roughness} \\
\hline
 & PSNR↑ & SSIM↑ & LPIPS↓ & PSNR↑ & SSIM↑ & LPIPS↓ & PSNR↑ & SSIM↑ & LPIPS↓ & PSNR↑ & SSIM↑ & LPIPS↓\\
NVDIFFREC & 30.674  & 0.954  & 0.105 & 26.198 & 0.945 & 0.110 & 12.317 & 0.627 & 0.370 & 20.498 & 0.898 & 0.238 \\
NVDIFFRECMC & 30.135 & 0.937 & 0.134 & 27.352 & 0.945 & 0.130 & 17.939 & 0.736 & 0.262 & 18.361 & 0.849 & 0.241 \\
Neural-PIL & 27.966 & 0.903 & 0.147 & 22.472 & 0.914 & 0.153 & 11.577 & 0.620 & 0.346 & 12.229 & 0.814 & 0.365 \\
NeILF & 30.172  & 0.968 & 0.057 & 24.027 & 0.949 & 0.105 & 13.918 & 0.668 & 0.272 & 12.232 & 0.795 & 0.313 \\
{\bf Ours} & {\bf 35.470} & {\bf 0.978} & {\bf 0.041} & {\bf 28.860} & {\bf 0.962} & {\bf 0.072} & {\bf 26.258} & {\bf 0.966} & {\bf 0.059} & {\bf 27.882} & {\bf 0.972} & {\bf 0.069}  \\
\hline
\end{tabular}
\end{table*}

\begin{figure*}
\centering
\includegraphics[width=0.95\linewidth]{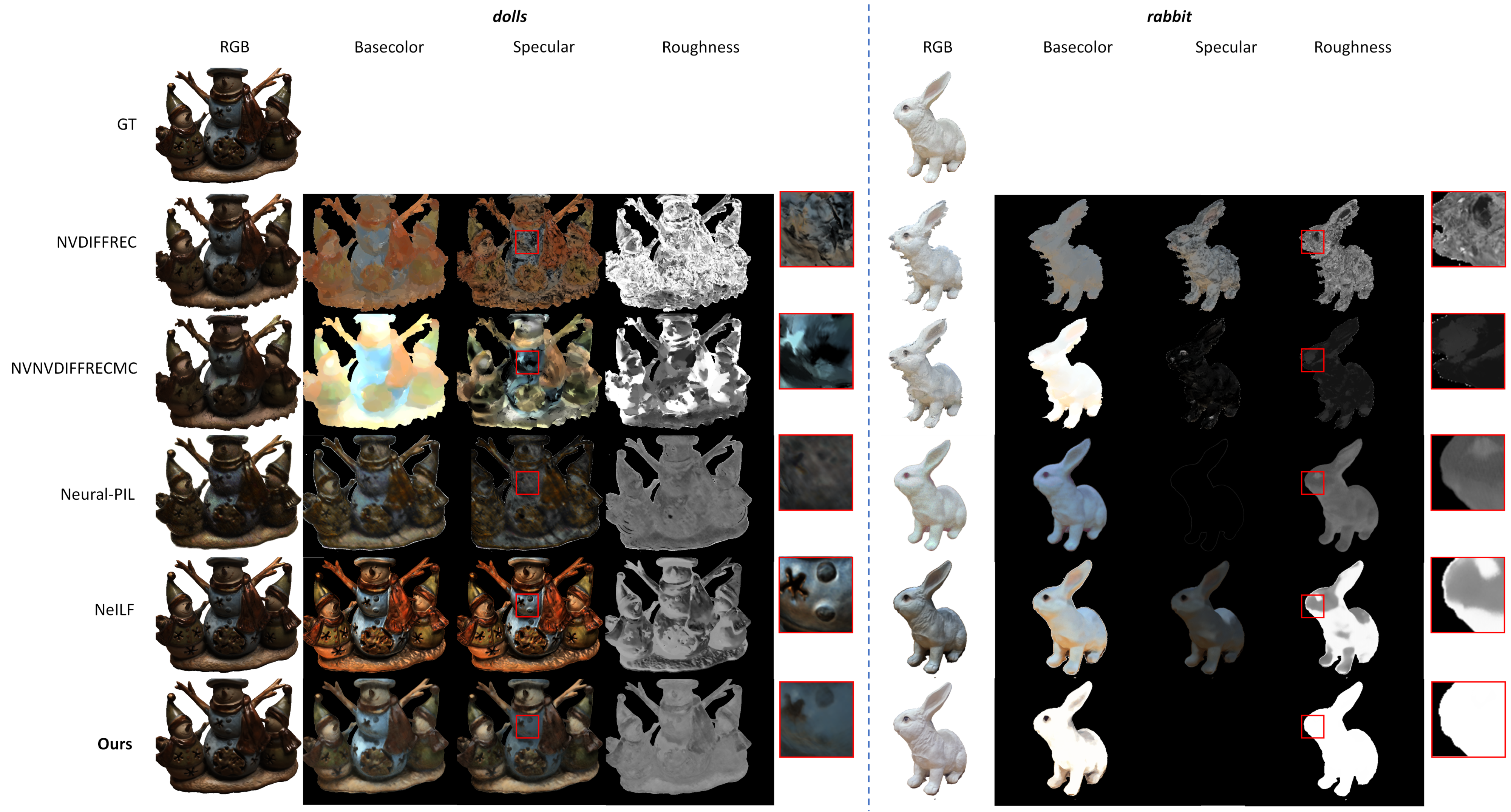}
\caption{Reconstruction and reflectance decomposition results on the real dataset. Our model achieves the best overall performance on the correctness and conciseness of material color, and our decomposition between individual BRDF attributes are also more accurate.}
\label{fig5}
\end{figure*}

\begin{table}
\caption{Reconstruction results on The Real Dataset.}
\label{tab:real}
\centering
\begin{tabular}{c|ccc}
\hline
\multicolumn{1}{c|}{} & \multicolumn{3}{c}{Reconstruction} \\
\hline
 & PSNR↑ & SSIM↑ & LPIPS↓\\
NVDIFFREC & 27.232  & 0.887  & 0.106 \\
NVDIFFRECMC & 25.238 & 0.872 & 0.132 \\
NeRFactor & 24.523  & 0.881 & 0.133 \\
Neural-PIL & 25.848  & 0.865 & 0.133 \\
NeILF & 27.005 & 0.923  & 0.070 \\
{\bf Ours} & {\bf 32.745} & {\bf 0.944} & {\bf 0.066} \\
\hline
\end{tabular}
\end{table}

\subsection{Reconstruction and Reflectance Decomposition}
\label{recon-exp}

We first evaluate the performance of our VQ-NeRF on both scene reconstruction and reflectance decomposition tasks. 
Table \ref{tab:nerfactor} and Fig. \ref{fig3} display the quantitative and qualitative results produced by baseline methods and ours on the CG data realised by previous methods. Due to these data lack specular and roughness references, we only perform comparison on the basecolor and final reconstruction results in Table \ref{tab:nerfactor} and Fig. \ref{fig3}. As a supplement, we show comprehensive evaluation results on CG data, including specular and roughness components, in Table \ref{tab:material} and Fig. \ref{fig4}. Furthermore, we perform reconstruction and decomposition on the real data, and show the comparison in Table \ref{tab:real} and Fig. \ref{fig5}. As there is no ground truth for materials in real data, the quantitative comparison only contains the reconstruction scores. Obviously, compare to baseline methods, our method demonstrates superior performance in terms of scene reconstruction and reflectance decomposition tasks, both on the CG dataset and the real dataset.

Specifically, the baseline methods NVDIFFREC and NVDIFFRECMC utilize rasterization-based rendering on explicit mesh, leading to noticeable artifacts such as stretched geometry and blurry textures in their outputs. For example, in the \textit{drums} scene depicted in Fig. \ref{fig3}, distinct noise is evident in their decomposed basecolor. Similarly, in the real-world \textit{dolls} scene shown in Fig. \ref{fig5}, these methods miss out on capturing intricate features like the eyes, noses, and decorations on the \textit{dolls} in their decomposed materials.
In contrast, although the implicit baselines do not suffer from these shortcomings, they come with their own limitations. NeRFactor and Neural-PIL, both relying on pre-trained networks for BRDF prediction, are heavily influenced by the distribution bias of the pre-training data. As a result, their decomposition results struggle with correctly predicting colors for diverse materials. This is evident in the \textit{metal-balls} scene illustrated in Fig. \ref{fig3}, where both methods fail to provide accurate color predictions. NeILF, on the other hand, lacks sufficient restrictions on illumination, leading to the absorption of lighting colors into the decomposed materials. This is evident in the \textit{rabbit} scene shown in Fig. \ref{fig5}. Additionally, since all these baselines primarily focus on continuous BRDF estimation, their decomposed materials tend to exhibit considerable noise, as observed in the \textit{kitchen} scene in Fig. \ref{fig4}. 

In contrast to these methods, our VQ-NeRF gets rid of all those limitations. By leveraging a mutually beneficial two-branch pipeline for reflectance decomposition, the material components produced by our method are free from artifacts and present the most accurate colors. Besides, with VQ discretization and dropout-based ranking strategy, our method enables compact predicted materials and suppresses the noise in the decomposed factors. Therefore, our method achieves reasonable scene reconstruction and reflectance decomposition, which in turn facilitates subsequent material editing and illumination editing.

\subsection{Material Editing}
\label{vqexp}

\begin{figure*}
\centering
\includegraphics[width=0.95\linewidth]{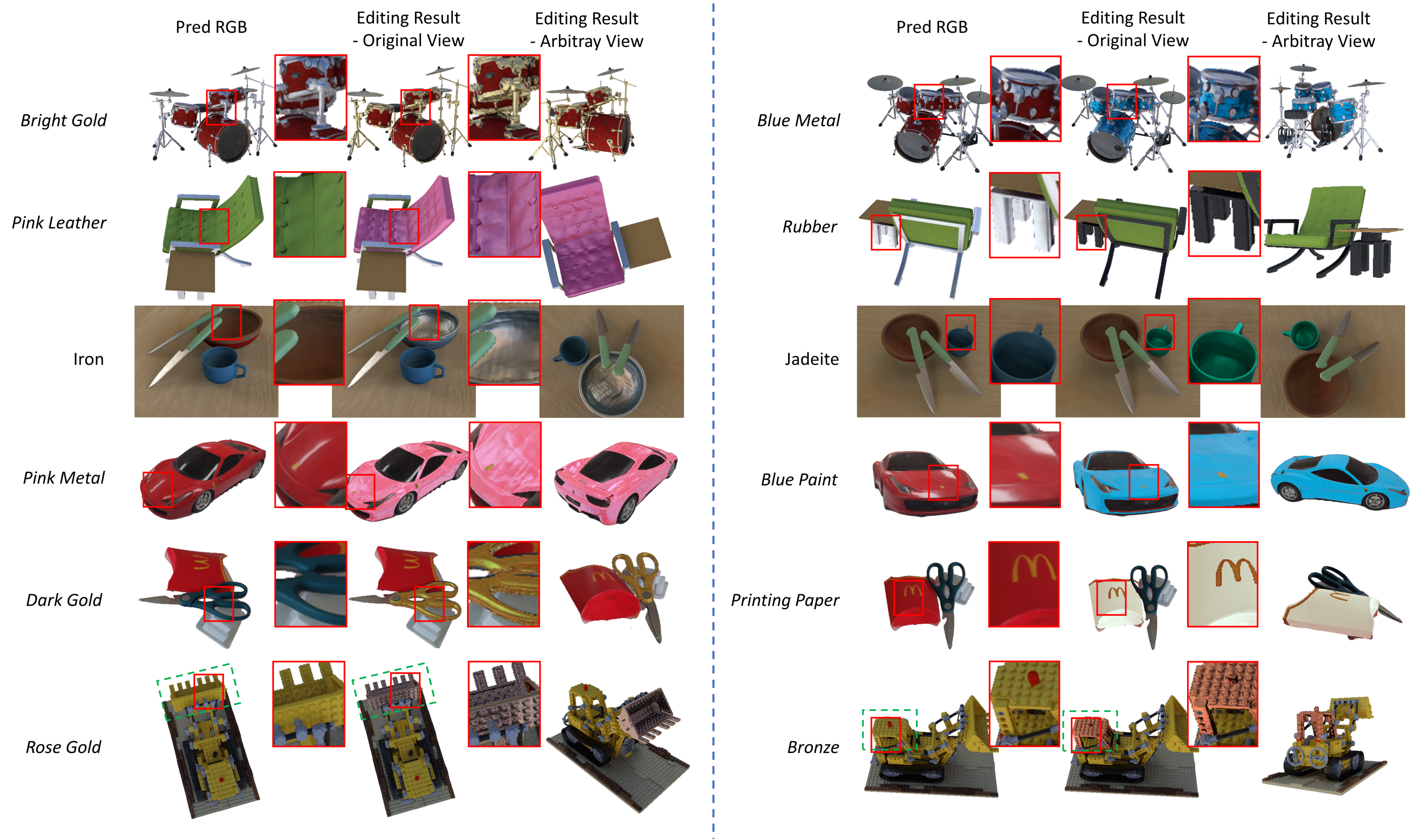}
\caption{Material editing results. With the help of our segmentation map, areas with the same material can be easily selected with just one or two mouse clicks and edited into a new material with specified BRDF parameters. Our model also supports local editing within bounding boxes, which are marked with green boxes in the \textit{lego} case.}
\label{fig10}
\end{figure*}

Since our VQ-NeRF is the first method to introduce discrete reflectance decomposition, it facilitates material selection and editing. The incorporation of the VQ mechanism not only suppresses decomposition noise but also supports the deduction of material segmentation maps in arbitrary views. Thanks to our dropout-based ranking strategy, the produced segmentation maps are compact and accurate. To achieve material editing, we first specify the desired material to be edited by selecting the corresponding area on the segmentation map. Subsequently, the selected material is replaced with new material with specified BRDF parameters, and the BRDF is applied to render the edited scene in arbitrary views. As shown in Fig. \ref{fig10}, all edits are made precisely in the corresponding areas, and obvious reflection changes such as highlight alternation are clearly visible in the edited model. In contrast to directly selecting the material to be edited in the segmentation map, other selection methods, such as bounding boxes, can also be combined with our segmentation map to support diverse local selection and editing, as shown in the \textit{lego} case. Please refer to our video for more details.

\begin{table}
\caption{Relighting Results on The CG Dataset. We use eight lighting probes provided by NVDIFFRECMC for quantitative evaluation.}
\label{tab:relit}
\centering
\begin{tabular}{c|ccc}
\hline
\multicolumn{1}{c|}{} & \multicolumn{3}{c}{Relighting} \\
\hline
 & PSNR↑ & SSIM↑ & LPIPS↓\\
NVDIFFREC & 23.385 & 0.892 & 0.082 \\
NVDIFFRECMC & 26.961 & 0.919 & 0.064 \\
NeRFactor & 24.850  & 0.915 & 0.067 \\
{\bf Ours} & {\bf 27.515} & {\bf 0.931} & {\bf 0.044} \\
\hline
\end{tabular}
\end{table}

\begin{figure*}
\centering
\includegraphics[width=0.95\linewidth]{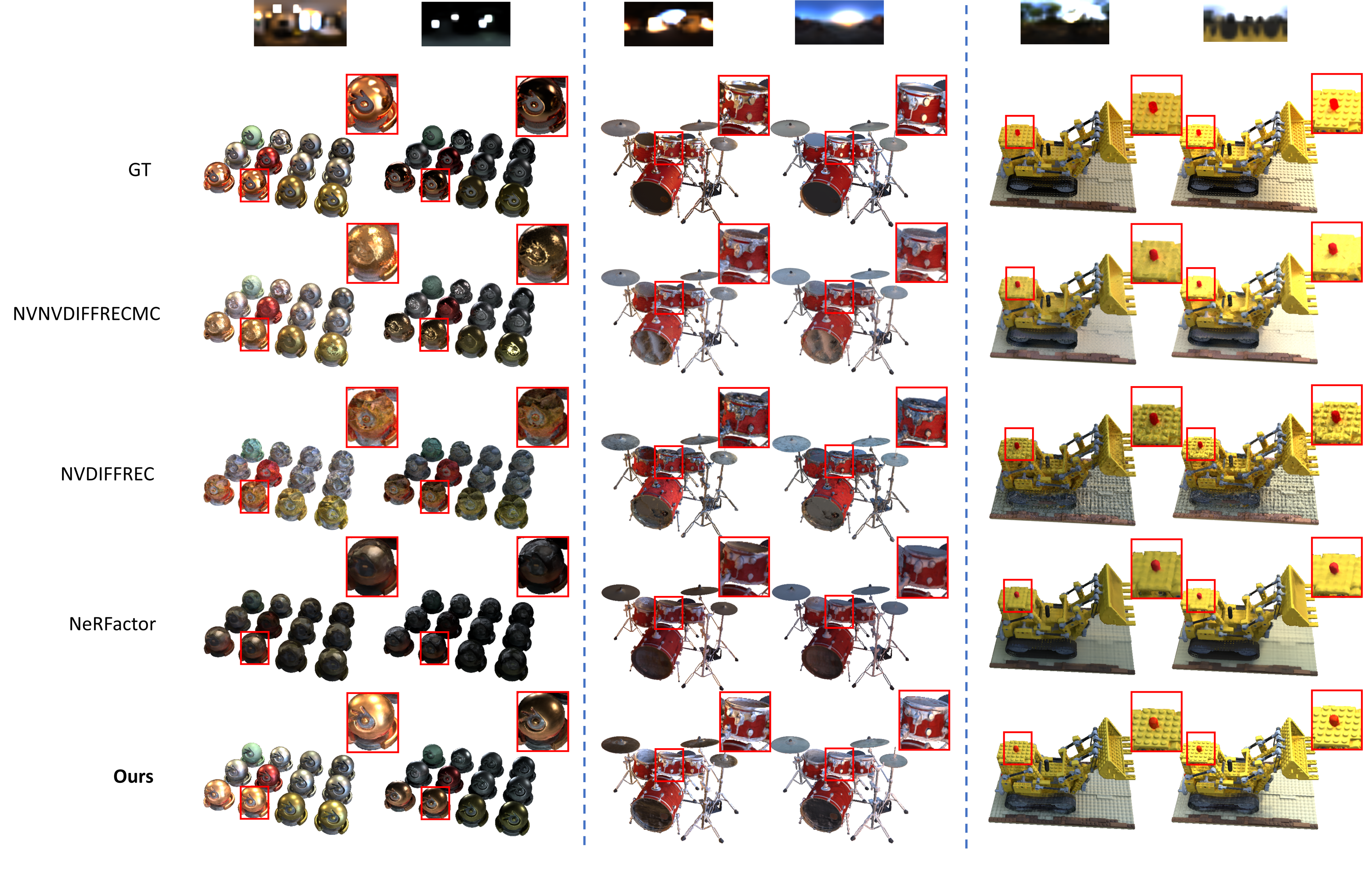}
\caption{Relighting results on the CG dataset. Evidently, our model produces realistic images with prominent highlights and clean appearance.}
\label{fig6}
\end{figure*}

\begin{figure*}
\centering
\includegraphics[width=0.95\linewidth]{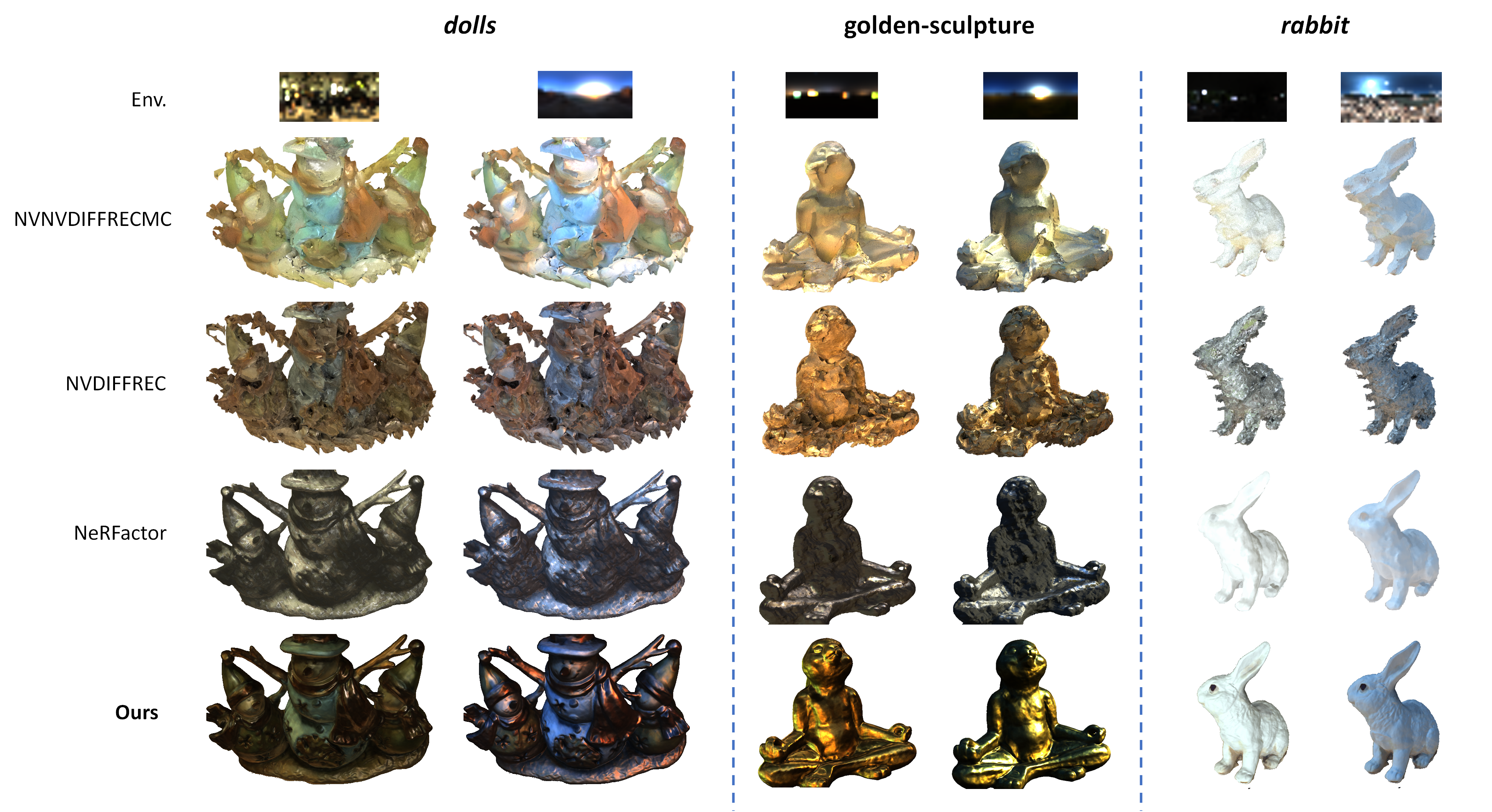}
\caption{Relighting results on the real dataset. In real-world scenes, our model can still produce realistic images with accurate colors and reflections.}
\label{fig7}
\end{figure*}

\subsection{Illumination Editing}

We further performed illumination editing on both CG data and real data, and compared our method with baseline methods that support relighting. Table \ref{tab:relit} and Fig. \ref{fig6} show the quantitative and visual relighting results on the CG data, respectively, while Fig. \ref{fig7} shows the results on the real data. Thanks to our high-quality reflectance decomposition, our method yields the most realistic relighted images with the highest metric scores compared to baseline methods.

Specifically, NVDIFFREC and NVDIFFRECMC struggled to produce clear and reasonable relighting scenes due to their explicit geometric representation and unreliable material modeling. Models produced by such methods contain a large number of rough surfaces caused by the extruded geometry, as shown in Fig. \ref{fig7}, which seriously degrade the quality of scene relighting. Unlike NVDIFFREC and NVDIFFRECMC, NeRFactor can produce plausible geometry thanks to its implicit scene representation. However, its reflectance decomposition module is based on a pre-trained BRDF prediction network, which limits its ability to accurately model materials and illumination. This is why NeRFactor fails to produce reasonable relighting colors in many cases, such as the \textit{metal-balls} scene in Fig. \ref{fig6}.

\begin{figure*}
\centering
\includegraphics[width=0.9\linewidth]{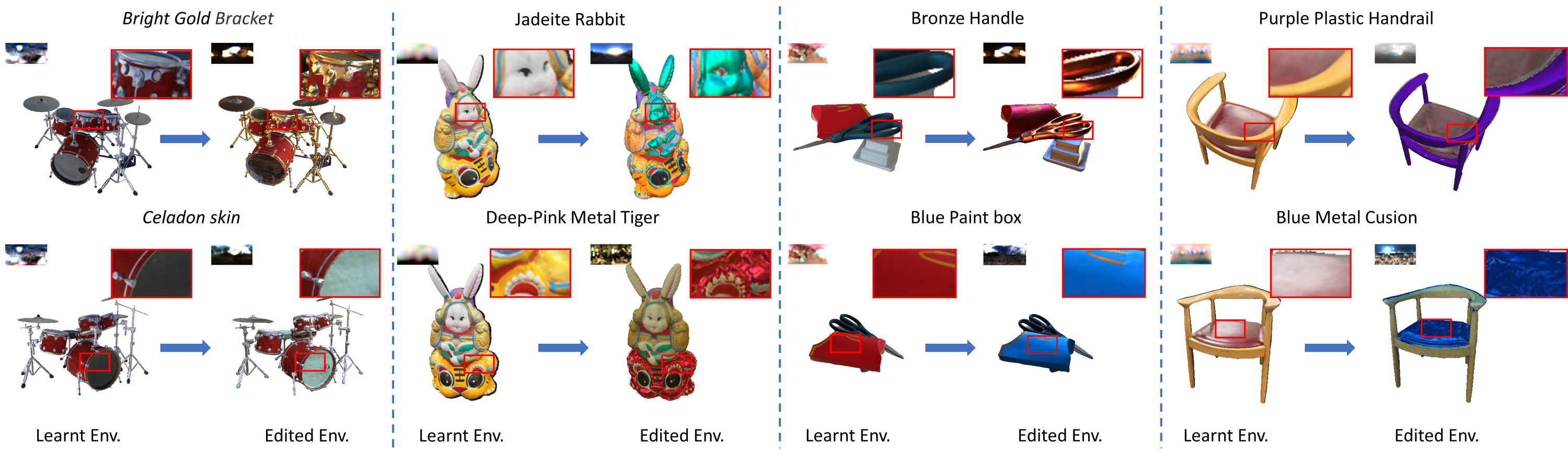}
\caption{Material-illumination joint editing results. The left-hand side shows the reconstructed image under the original scene lighting, and the right-hand side shows the relighted, material-modified images. Obviously, our editing results in different cases remain highly realistic.}
\label{fig11}
\end{figure*}

In contrast, our VQ-NeRF is able to generate reasonable relighting results, exhibiting proper color and obvious highlight reflections under different environmental lighting, as shown in the \textit{drums} and \textit{lego} scenarios of Fig. \ref{fig6}. Moreover, since our VQ-NeRF enables efficient material editing and scene relighting, our method also supports simultaneous material and lighting editing. As shown in Fig. \ref{fig11}, the left column shows the reconstructed result with the original lighting, while the right column shows the edited result with novel material and lighting. 

\subsection{Ablation Study}

\begin{figure}
\centering
\includegraphics[width=0.95\linewidth]{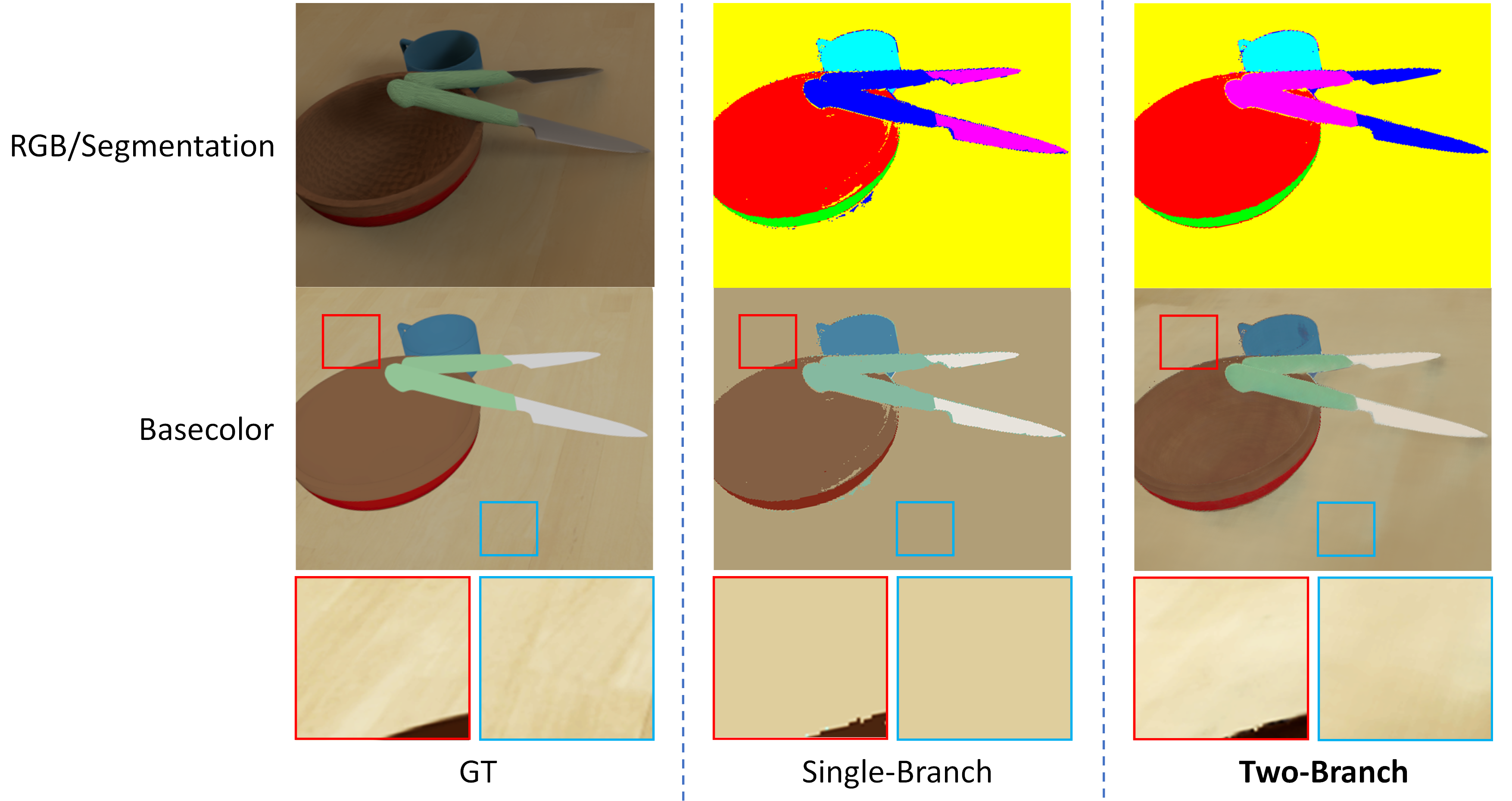}
\caption{Comparison of single-branch and two-branch frameworks. The contrast and saturation of local regions within the color boxes are adjusted for a better comparison. The single-branch model produces completely flattened basecolor and cannot reconstruct subtle variations within a single material. In contrast, our two-branch framework allows for small variations within the same type of material, resulting in better reconstruction of details in the scene.}
\label{fig18}
\end{figure}

\subsubsection{Two-Branch vs. Single-Branch}

In our two-branch framework, the continuous branch outputs  decomposition, while the discrete branch yields the segmentation map for material clustering. To validate the effectiveness of our two-branch design, we compare it with a single-branch option. Specifically, we use the discrete branch to simultaneously learn reflectance decomposition and VQ clustering, and as a result, it outputs both a segmentation map and BRDF factors after VQ discretization. As shown in Fig. \ref{fig18}, both single-branch and two-branch options can generate the correct segmentation map. However, the single-branch model produces completely flattened basecolor and cannot reconstruct subtle variations within a single material, such as the wooden texture, because the reflectance decomposition is strictly constrained by the VQ clustering. In contrast, in our two-branch framework, the VQ clustering constraint imposed by the discrete branch is a soft constraint, which facilitates the continuous branch in performing more discrete reflectance decomposition while still allowing for small variations within the same type of material to better reconstruct. This illustrates the essential role of the two-branch design in our method. 

\begin{figure}
\centering
\includegraphics[width=0.95\linewidth]{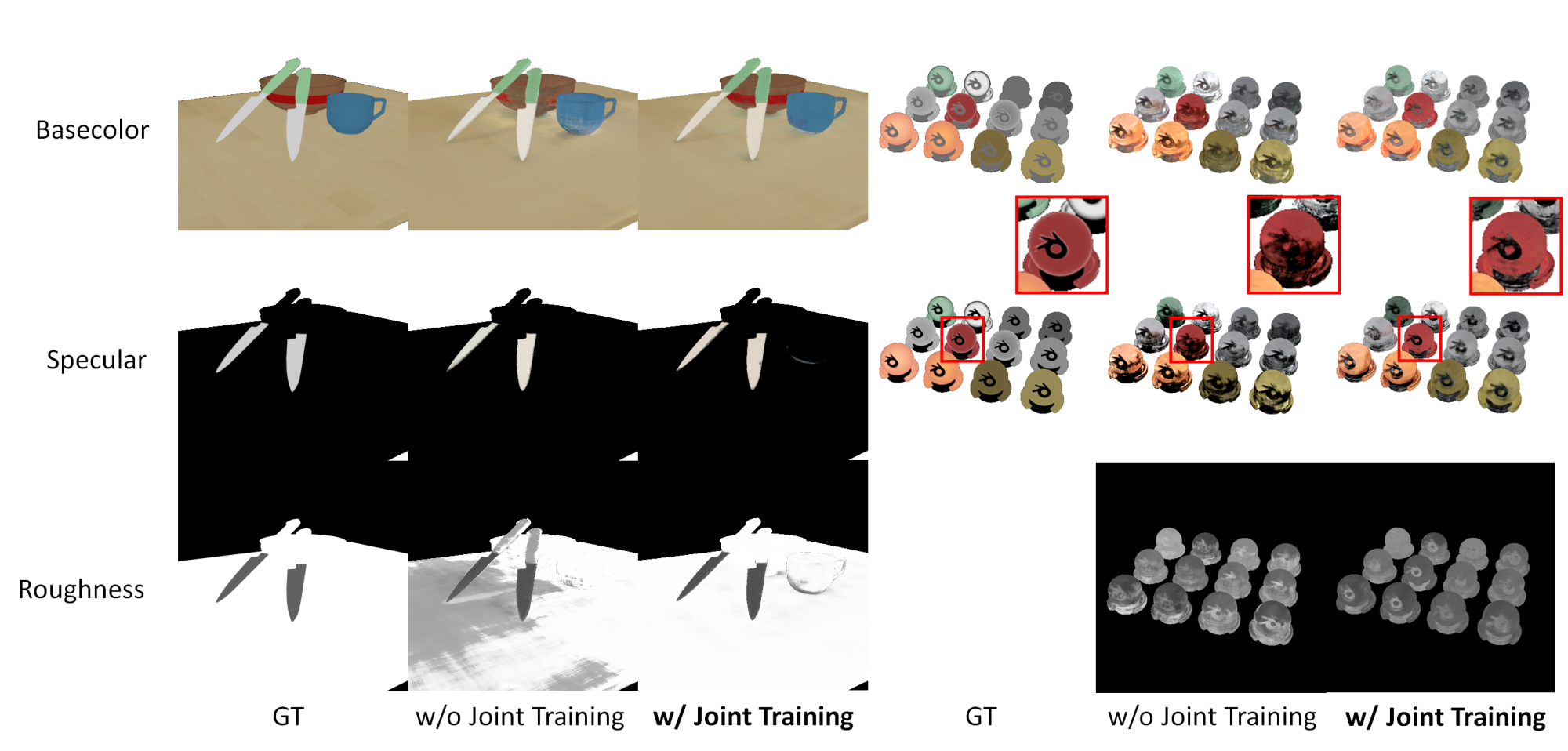}
\caption{During the two-branch joint training, VQ clustering compacts the latent distribution of material vectors, reducing decomposition noise.}
\label{fig13}
\end{figure}

\begin{figure}
\centering
\includegraphics[width=0.95\linewidth]{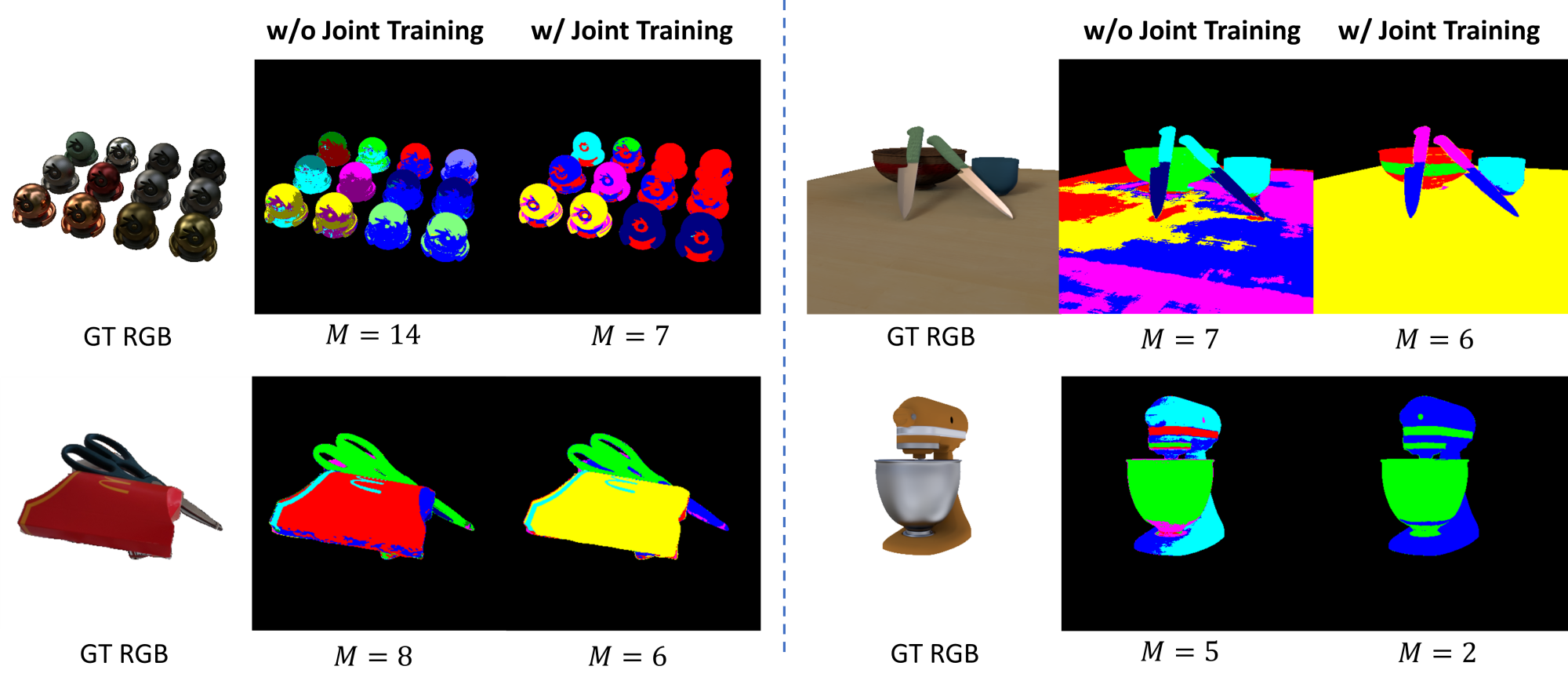}
\caption{When adopting joint training, VQ discretization produces accurate segmentation with less redundancy.}
\label{fig17}
\end{figure}

\begin{table*}
\caption{Ablation Study. Two-branch joint training can improve the decomposition accuracy.}
\label{tab:wovq}
\centering
\begin{tabular}{c|ccc|ccc|ccc}
\hline
\multicolumn{1}{c|}{} & \multicolumn{3}{c|}{Basecolor} & \multicolumn{3}{c|}{Specular} & \multicolumn{3}{c}{Roughness} \\
\hline
 & PSNR↑ & SSIM↑ & LPIPS↓ & PSNR↑ & SSIM↑ & LPIPS↓ & PSNR↑ & SSIM↑ & LPIPS↓ \\
w/o joint training & 27.441 & 0.953 & 0.078 & 24.918 & 0.963 & {\bf 0.058} & 22.539 & 0.926 & 0.172 \\
Ours (w/ joint training) & {\bf 28.860} & {\bf 0.962} & {\bf 0.072} & {\bf 26.258} & {\bf 0.966} & 0.059 & {\bf 27.882} & {\bf 0.972} & {\bf 0.069} \\
\hline
\end{tabular}
\end{table*}

\subsubsection{Ablation on Joint Training}
\label{ab-vq}
To investigate the effectiveness of the joint training strategy of the two branches, we conducted an experiment by replacing the joint training with a separate training strategy. Specifically, we first trained the continuous branch independently of the discrete branch. Then, we fixed the continuous branch and trained the discrete branch. The quantitative results are reported in Table \ref{tab:wovq}, and the visualization results are shown in Figs. \ref{fig13} and \ref{fig17}. Compared to the separate training strategy, the joint training strategy allows the continuous and discrete branches to benefit from each other during training. Specifically, our method with joint training can effectively suppress the noise in the predicted decomposed material, as shown in Fig. \ref{fig13}. Meanwhile, the material components produced by our method are clean and reasonable. In contrast, the method without joint training may produce confused materials, such as the knife handle and cup in the \textit{kitchen} scene in Fig. \ref{fig17}.

\begin{table}
\caption{Material Segmentation Results on The CG Dataset. The values of the three metrics under micro average are the same, so we only present Micro-F1 for comparison.}
\label{tab:cluster}
\centering
\begin{tabular}{c|cccc}
\hline
\multicolumn{1}{c|}{} & \multicolumn{4}{c}{Material Clustering} \\
\hline
 & Micro-F1↑ & Macro-F1↑ & Macro-P↑ & Macro-R↑\\
Meanshift-0.2 & 0.747 & 0.350 & 0.357 & 0.384 \\
Meanshift-0.3 & 0.629 & 0.268 & 0.296 & 0.303 \\
Meanshift-0.5 & 0.503  & 0.127 & 0.146 & 0.173 \\
{\bf Ours} & {\bf 0.821} & {\bf 0.421} & {\bf 0.405} & {\bf 0.449} \\
\hline
\end{tabular}
\end{table}

\begin{figure*}
\centering
\includegraphics[width=0.95\linewidth]{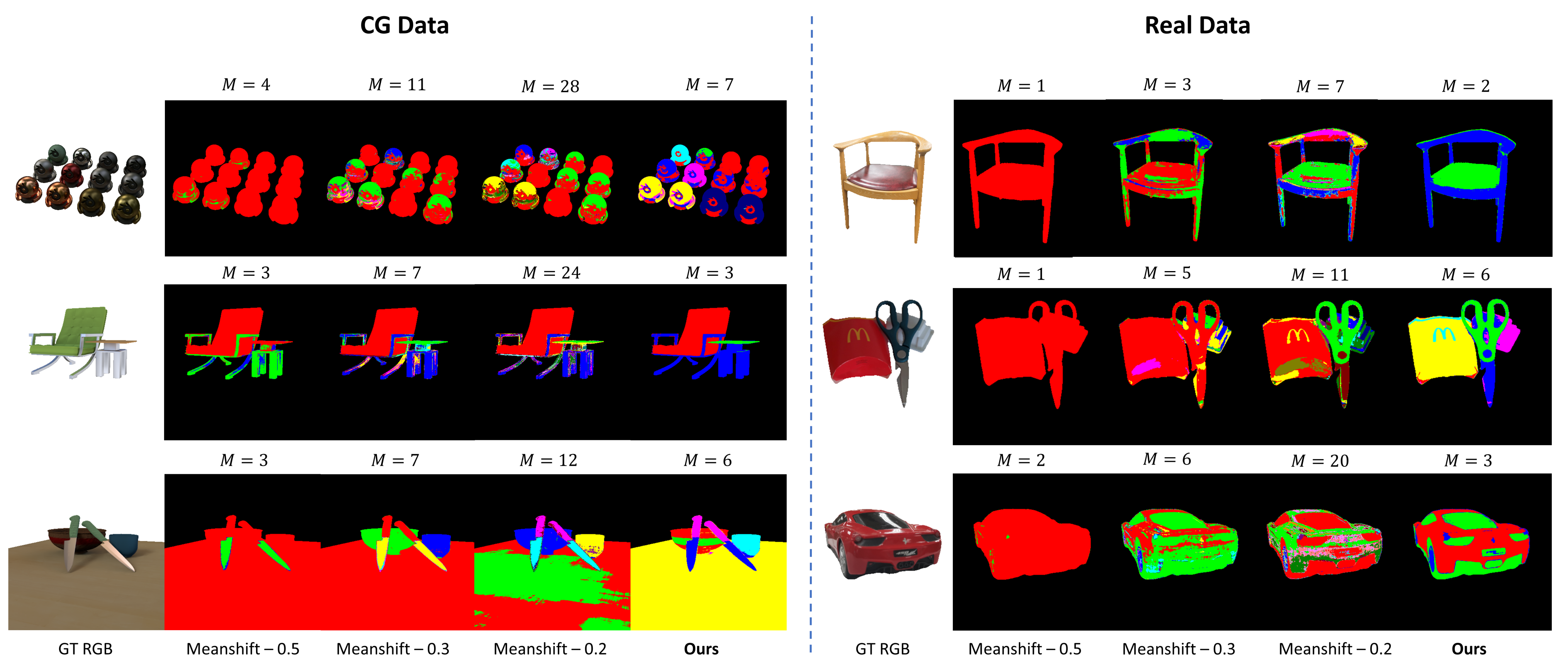}
\caption{Material segmentation results. $M$ refers to the predicted number of scene materials. The left column shows the results on CG data, and the right column shows the results on real data. Evidently, our strategy produces the most accurate and least redundant segmentation on both CG and real data, demonstrating the superiority of our VQ clustering in discrete branch.}
\label{fig8}
\end{figure*}

\subsubsection{VQ Clustering vs. Classical Clustering}
\label{segexp}
To investigate the superiority of our VQ clustering, we conducted an experiment by replacing it with an intuitive clustering method. Specifically, we applied the \textit{meanshift} clustering with three different bandwidths (0.5, 0.3, and 0.2) on our continuous branch for discrete material segmentation. As shown in Table \ref{tab:cluster} and Fig. \ref{fig8}, both quantitative and qualitative results demonstrate that the material segmentation produced by our VQ mechanism is more accurate and effective compared to the classical \textit{meanshift} clustering method. For example, in the \textit{metal-balls} scenario of Fig. \ref{fig8}, the \textit{meanshift} clustering with different bandwidths fails to deduce reasonable and correct segmentation maps from only the continuous branch, while our VQ clustering in the discrete branch enables us to identify all the materials and infer an accurate material segmentation map. This advantage is attributed to learning clustering jointly with decomposition.

\subsubsection{Ablation on Dropout-based Ranking}

\begin{figure}
\centering
\includegraphics[width=0.95\linewidth]{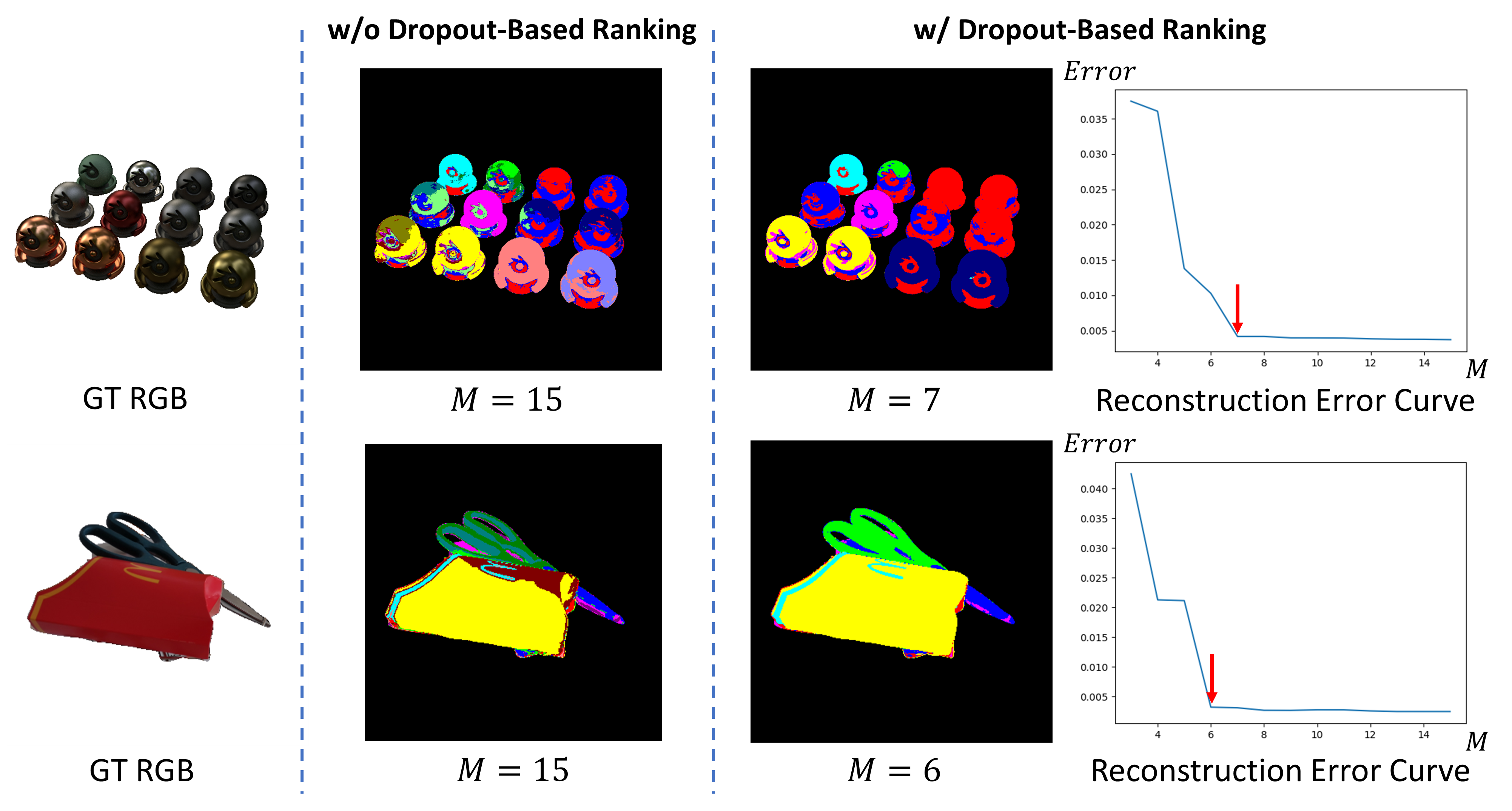}
\caption{Ablation results concerning the dropout-based ranking strategy, with the fourth column showing the reconstruction error curves. The horizontal axis of the curves represents the number of used materials $M$, while the vertical axis represents the reconstruction error. As depicted, the reconstruction error decreases as more materials are used. However, as shown by the results in the second column, using too many materials may result in redundant segmentation. Leveraging the dropout-based ranking strategy, we achieve a balance between reconstruction error and material redundancy, as demonstrated in the third column.}
\label{fig12}
\end{figure}

To illustrate the effectiveness of our dropout-based ranking strategy, we further compare our full method with the implementation without this strategy. Fig. \ref{fig12} shows the visual comparison on discrete material segmentation. Here, in the implementation without the dropout-based ranking strategy, since the length of the VQ codebook cannot be automatically detected without our dropout-based ranking strategy, we set a constant $M=15$ as the codebook length and show the segmentation map inferred from all codewords. Unlike the implementation without ranking strategy, our full method is able to rank codewords in importance and automatically determine the length of the VQ codebook, i.e., the number of materials. Therefore, our full method achieves to discard redundant codewords and generate compact material segmentation maps, which greatly facilitates the subsequent material selection and editing process. Moreover, to prove the rationality of the length selection, we also show the reconstruction error curve as the number of materials changes in Fig. \ref{fig12}. As shown in the figure, the reconstruction error decreases as more materials are used. However, using too many materials may result in redundant segmentation. By introducing the dropout-based ranking strategy, our full method is able to select the most appropriate number of materials to balance reconstruction error and material redundancy.

\begin{figure}
\centering
\includegraphics[width=0.9\linewidth]{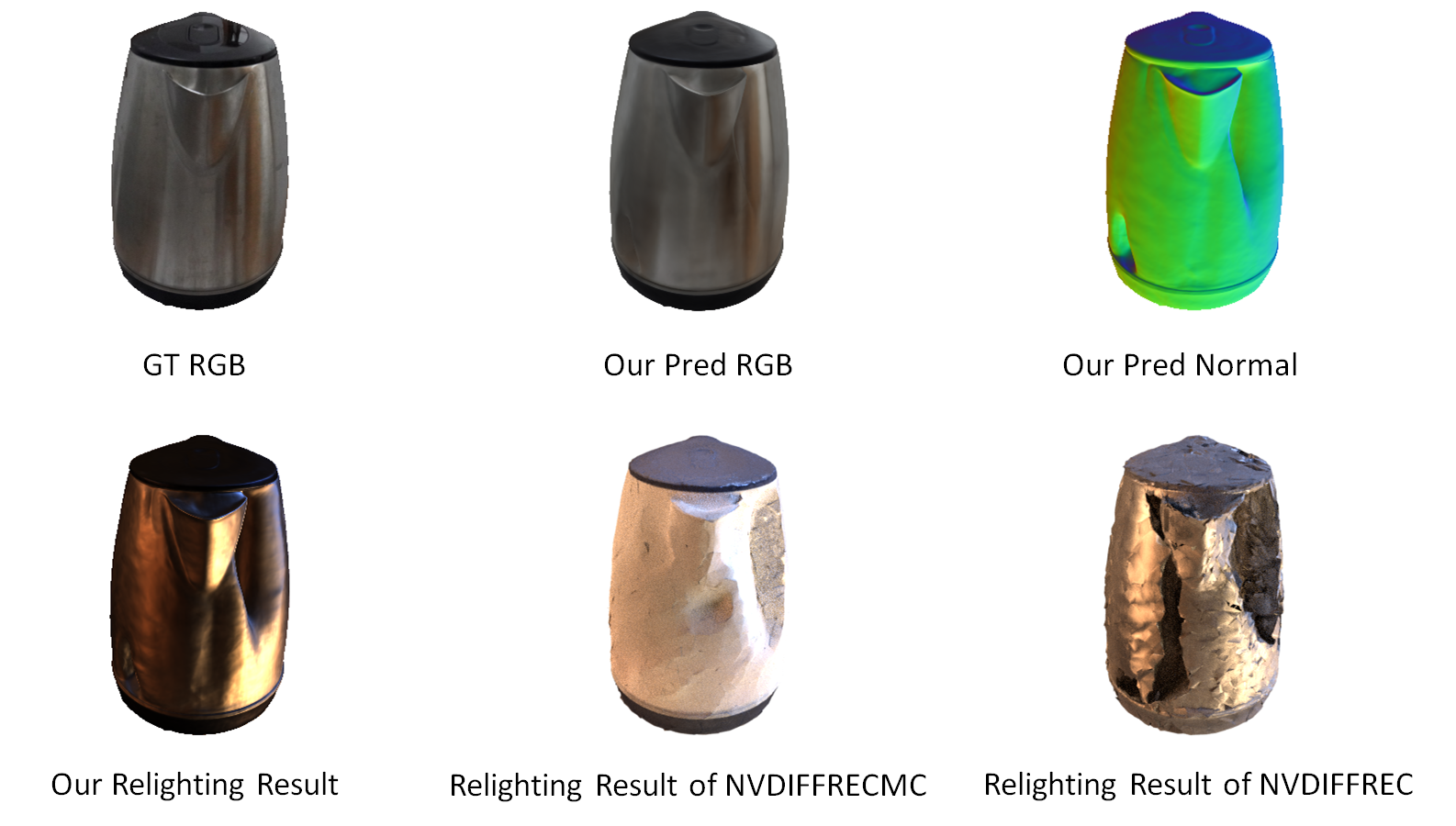}
\caption{Like other reflectance decomposition methods, inaccurate shapes produced by the geometry reconstruction method may result in relighting failures in our method.}
\label{fig15}
\end{figure}

\section{Conclusion}

In this paper, we propose VQ-NeRF, a VQ-based two-branch neural reflectance field for reflectance decomposition, material editing, and scene relighting. Unlike existing methods generate continuous material components, our VQ-NeRF introduce a discrete branch in addition to the continuous branch to produce discrete materials and deduce segmentation maps for facilitating material selection and editing. Meanwhile, we employ a dropout-based ranking strategy to eliminate material redundancy and automatically determine the number of scene materials. Moreover, we adopt a two-branch joint training strategy to encourage mutual benefit between the continuous and discrete branches and suppress the noise predicted in material components. Extensive experiments on both CG and real data demonstrate the superior performance of our VQ-NeRF in scene reconstruction, reflectance decomposition, material editing, and scene relighting tasks.

\noindent\textbf{Limitations.} In some cases, our method may fail to generate correct relighting or material editing results when the geometry reconstruction method (e.g., NeuS) cannot correctly model the geometric surface, as shown in Fig. \ref{fig15}. However, this limitation can potentially be overcome by developing more advanced neural implicit 3D representations or by jointly learning geometric reconstruction and reflectance decomposition. We leave these as future directions for our work.

\section*{Acknowledgments}
The work described in this paper was fully supported by a GRF grant from the Research Grants Council (RGC) of the Hong Kong Special Administrative Region, China [Project No. CityU 11208123]. 

\bibliographystyle{IEEEtran}
\bibliography{egbib}

\end{document}


\title{Supplementary Material for VQ-NeRF}

\maketitle

In the following, we supplement the paper with additional results and details, including:
\begin{itemize}
\item More details about our geometry extraction.
\item More details about our two-branch network.
\item More information about our data collection.
\item More experiments on our model, including ablation of loss functions and visualization of model optimization.
\item A supplementary video.
\end{itemize}

\section{Geometry Extraction}

In our implementation, we employ the NeuS \cite{wang2021neus1} for geometry reconstruction. In Neus, the geometry of a scene is modeled by an SDF $f$. Denote $\boldsymbol{r}(t)=\boldsymbol{o}+t\boldsymbol{d}$ as the spatial points sampled on the camera ray emitted from the origin $\boldsymbol{o}$ in the direction $\boldsymbol{d}$. The surface point $\boldsymbol{p}$ of ray $\boldsymbol{r}$ can be computed by:
\begin{equation}
    \begin{cases}
        \boldsymbol{p}(\boldsymbol{r})=\int_{t_{n}}^{t_{f}}T(t)\rho(f(\boldsymbol{r}(t)),t)\boldsymbol{r}(t){dt}, \\   
        T(t)=\exp(-\int_{t_{n}}^{t}\rho(f(\boldsymbol{r}(u)),u){du}),
    \end{cases}
\end{equation}
where $t_n$ and $t_f$ represent the bounds of near and far sampling. $f(\boldsymbol{r}(t))$ represents the value of the SDF $f$ at point  $\boldsymbol{r}(t)$. $\rho(f(\boldsymbol{r}(t)),t)$ is the opaque density. Additionally, we follow the implementation of NeuS to compute the surface normal $\boldsymbol{N}(\boldsymbol{p})$:
\begin{equation}
    \boldsymbol{\overline{N}}(\boldsymbol{p})=\int_{t_{n}}^{t_{f}}T(t)\rho(f(\boldsymbol{r}(t)),t){\nabla_{\boldsymbol{r}(t)}f}{dt},
\end{equation}
where $\nabla_{\boldsymbol{r}(t)}f$ is the gradient of $f$ with respect to $\boldsymbol{r}(t)$. Then, we normalize $\boldsymbol{\overline{N}}(\boldsymbol{p})$ by $\boldsymbol{N}(\boldsymbol{p})=\frac{\boldsymbol{\overline{N}}(\boldsymbol{p})}{||\boldsymbol{\overline{N}}(\boldsymbol{p})||}$ to obtain the final result.

\section{Network Structure}

\subsubsection{Encoder Structure}
\begin{figure}[h]
\centering
\includegraphics[width=0.9\linewidth]{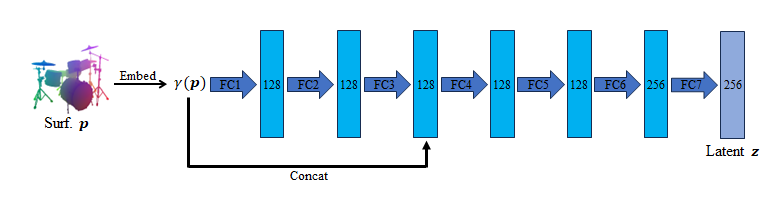}
\caption{The architecture of the encoder.}
\label{fig:enc}
\end{figure}

The structure of the encoder $f_e$ is illustrated in Fig. \ref{fig:enc}. The dark blue arrows indicate the fully-connected layers. The light blue squares represent the latent vectors. The numbers enclosed within the squares denote the vector dimensions. The input to the encoder is the surface point $\boldsymbol{p}$. $\boldsymbol{p}$ is first embedded by the positional encoding $\gamma$. Then, the embedded $\gamma(\boldsymbol{p})$ is processed through seven fully-connected layers to generate the latent material vector $\boldsymbol{z}$. A skip connection is established between the embedded inputs and the output vectors of the third layer.

\subsubsection{Decoder Structure}

\begin{figure}[h]
\centering
\includegraphics[width=0.9\linewidth]{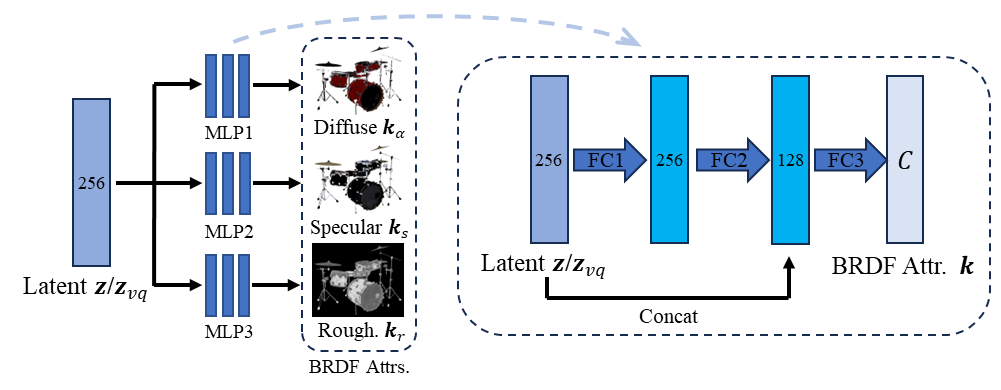}
\caption{The architecture of the decoders.}
\label{fig:dec}
\end{figure}

We employ similar architecture for the decoders $f_d^c$ and $f_d^d$, as shown in Fig. \ref{fig:dec}. Both $f_d^c$ and $f_d^d$ consist of three MLPs for predicting different BRDF attributes (diffuse, specular, or roughness). In each MLP, the latent material vector $\boldsymbol{z}$ or the quantized codeword $\boldsymbol{z}_{vq}$ is processed through three fully-connected layers to predict the BRDF attribute with $C$ channels. A skip connection is established between the inputs and the output vectors of the second layer. In the specific implementation, $f_d^c$ uses the simplified Disney BRDF \cite{yao2022neilf1} for rigorous constraints. Therefore, its outputs require a linear transformation before rendering, as described in Sec. III-B of the paper.

\section{Data Collection}

Due to some setting differences in source files, some scenes in the data released by NeRF and NeRFactor do not have ground truth of specular and roughness. To comprehensively evaluate our decomposition accuracy, we additionally collect three CG scenes containing the ground-truth of all BRDF attributes, and make our own dataset.

The geometry models, textures and illuminations used in these scenes are all collected online. Our data contains a variety of materials such as wood, metal, plastic, and fabric, allowing exhaustive evaluation of material decomposition. All of the scenes are lighted by HDR illumination maps, which are in the same format as the public data provided by NeRFactor. For rendering, we use the CYCLES engine of the Blender software, and employ the same camera poses used in the NeRF and NeRFactor datasets. We keep the same rendering settings as NeRFactor and use their released scripts to generate our data.

\begin{figure}[t]
\centering
\includegraphics[width=0.9\linewidth]{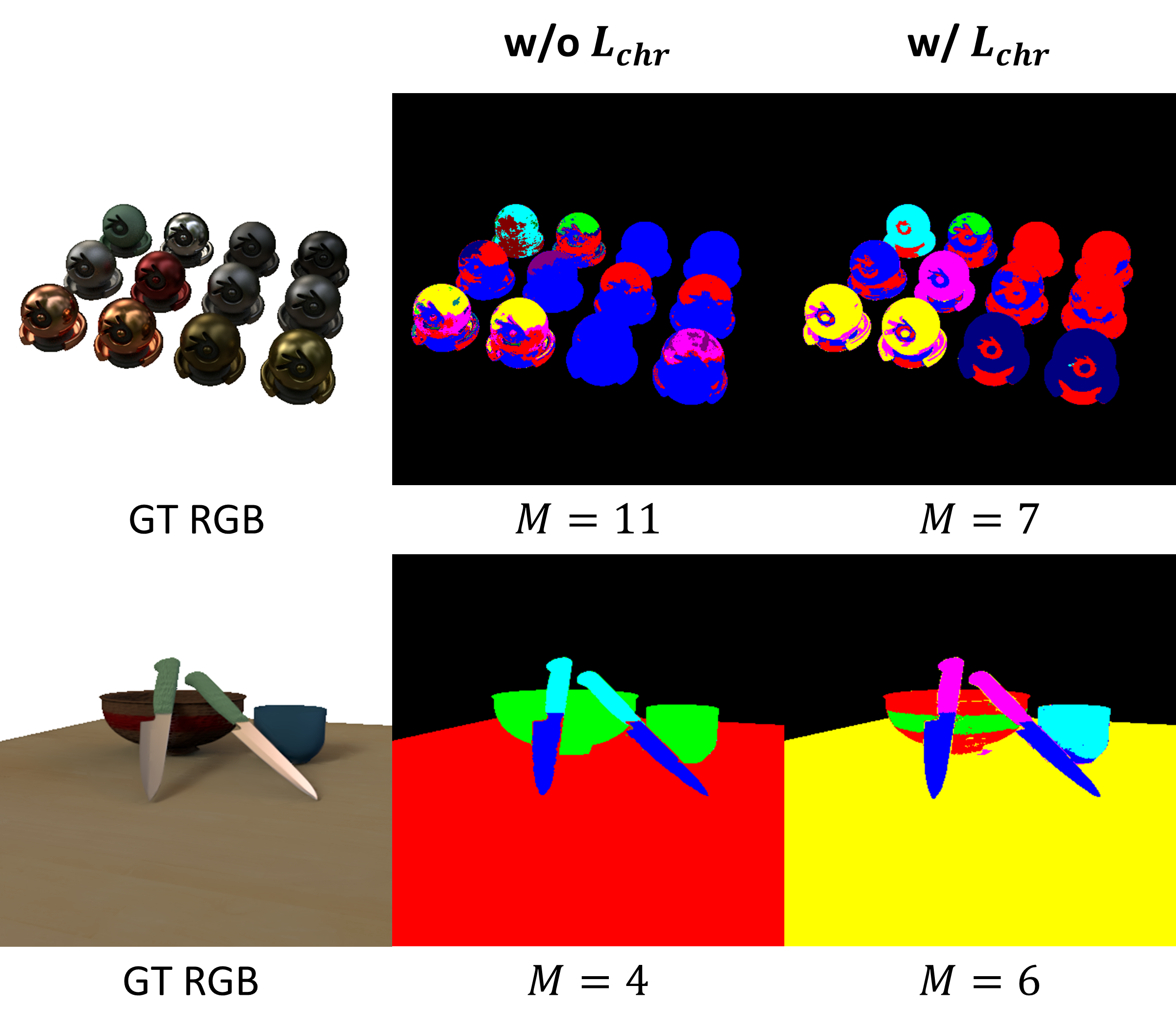}
\caption{Ablation on chromaticity loss. The introduction of chromaticity loss improves the segmentation accuracy of surface points with dark colors.}
\label{fig:lchr}
\end{figure}

\begin{table}[h]
\caption{Ablation on chromaticity loss.}
\label{tab:lchr}
\centering
\begin{tabular}{c|cccc}
\hline
\multicolumn{1}{c|}{} & \multicolumn{4}{c}{Material Clustering} \\
\hline
 & Micro-F1↑ & Macro-F1↑ & Macro-P↑ & Macro-R↑\\
w/o $L_{chr}$ & 0.755 & 0.374 & 0.376 & 0.406 \\
Ours (w/ $L_{chr}$) & {\bf 0.821} & {\bf 0.421} & {\bf 0.405} & {\bf 0.449} \\
\hline
\end{tabular}
\end{table}

\begin{table*}[h]
\caption{Ablation study. Although the Lambertian loss slightly reduces the accuracy of basecolor and roughness, it significantly enhances the decomposition of the specular attribute.}
\label{tab:llam}
\centering
\begin{tabular}{c|ccc|ccc|ccc}
\hline
\multicolumn{1}{c|}{} & \multicolumn{3}{c|}{Basecolor} & \multicolumn{3}{c|}{Specular} & \multicolumn{3}{c}{Roughness} \\
\hline
 & PSNR↑ & SSIM↑ & LPIPS↓ & PSNR↑ & SSIM↑ & LPIPS↓ & PSNR↑ & SSIM↑ & LPIPS↓ \\
w/o $L_{lam}$ & {\bf 29.914} & {\bf 0.967} & {\bf 0.059} & 13.676 & 0.639 & 0.272 & {\bf 30.386} & {\bf 0.979} & {\bf 0.043} \\
Ours (w/ $L_{lam}$) & 28.860 & 0.962 & 0.072 & {\bf 26.258} & {\bf 0.966} & {\bf 0.059} & 27.882 & 0.972 & 0.069 \\
\hline
\end{tabular}
\end{table*}

\section{Supplementary Experiments}

\subsection{Ablation on Loss Functions}

To demonstrate the necessity of adopting $L_{chr}$, $L_{lam}$ and $L_{sm}$ in model optimization, we conduct more experiments to evaluate their impact on material decomposition and segmentation.

\begin{figure*}[b]
\centering
\includegraphics[width=0.95\linewidth]{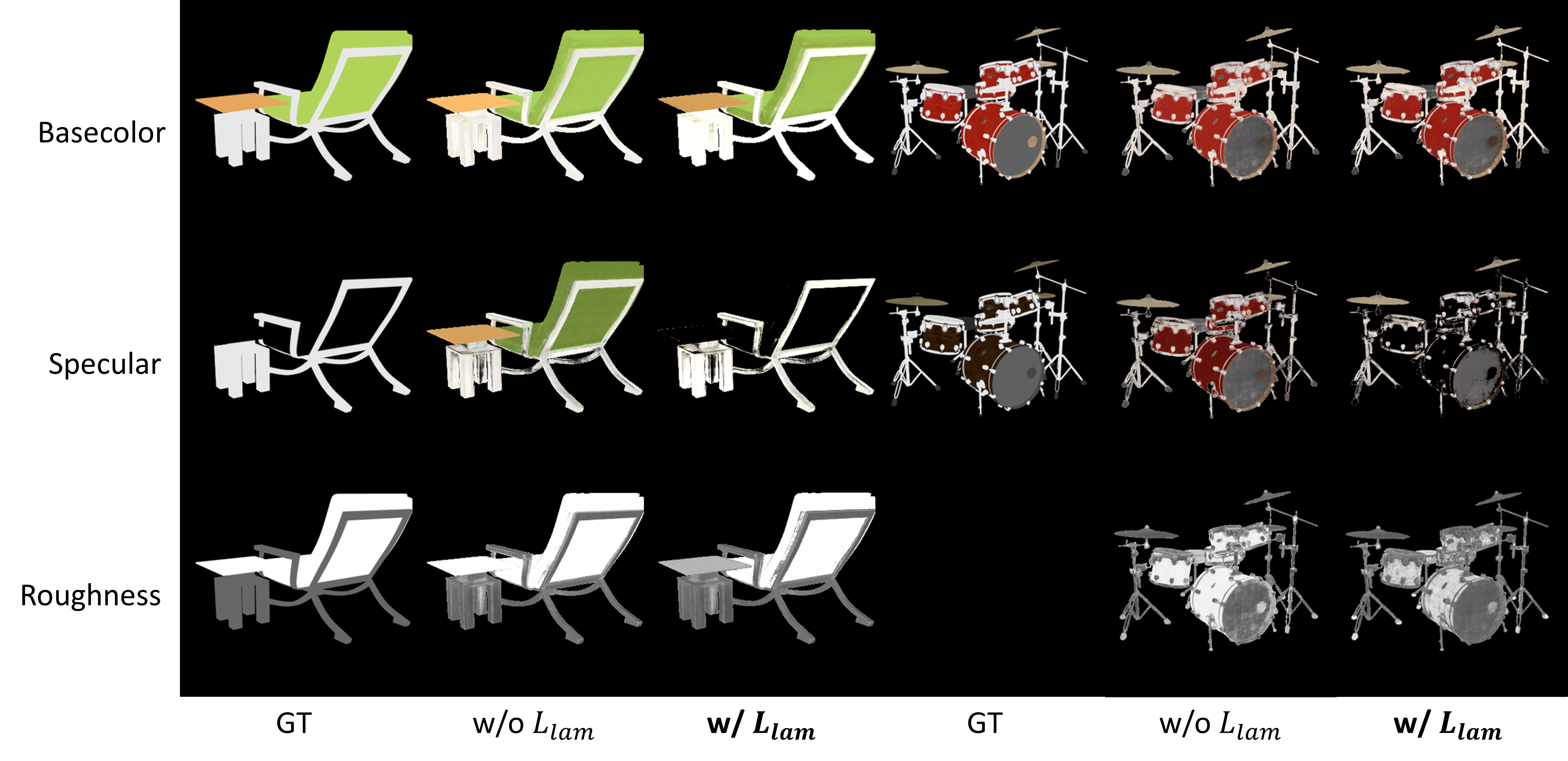}
\caption{Ablation on Lambertian loss. The incorporation of Lambertian loss significantly improves the decomposition accuracy of specular attributes.}
\label{fig:llam}
\end{figure*}

\subsubsection{Ablation on $L_{chr}$}

Our model works on data captured under arbitrary lighting, and in some scenes, the luminance of the appearance color varies greatly across different surface points. Therefore, those surface points with dark colors only have a small loss when computing the reconstruction loss in RGB space. This presents challenges for the model optimization on those dark points. Such problem is severe in our VQ segmentation, leading to confusion between different materials with dark colors. To address this issue, we introduce a chromaticity loss $L_{chr}$ in the discrete branch. $L_{chr}$ computes the reconstruction loss in chromaticity space, where all colors are normalized to the same lightness 1. Therefore, the computed loss is independent of the luminance of apperance color, so surface points with dark colors can be strongly constrained, leading to better convergence.

We compare our method to a model trained without $L_{chr}$. Table \ref{tab:lchr} and Fig. \ref{fig:lchr} show the quantitative and qualitative results. Apparently, the chromaticity loss improves the segmentation accuracy on surface points with dark colors, such as the matte parts of red balls and black-green balls in the \textit{metal-balls} scene in Fig. \ref{fig:lchr}.

\subsubsection{Ablation on $L_{lam}$}

Due to the ambiguity of inverse rendering, different combinations of BRDF attributes can render to the same appearance. A typical example is that a specular material with high roughness can exhibit a similar appearance to a diffuse material. To introduce more constraints to resolve this ambiguity, we propose a Lambertian loss that predicts surface points with high roughness to have low specular, which is consistent with real material relations.

We compare our method to a model trained without $L_{lam}$. Table \ref{tab:llam} and Fig. \ref{fig:llam} give the quantitative and qualitative results. As shown, although the incorporation of $L_{lam}$ slightly reduces the decomposition accuracy of the basecolor and roughness, it significantly improves the decomposition quality of the specular attribute.

\begin{figure}[t]
\centering
\includegraphics[width=0.99\linewidth]{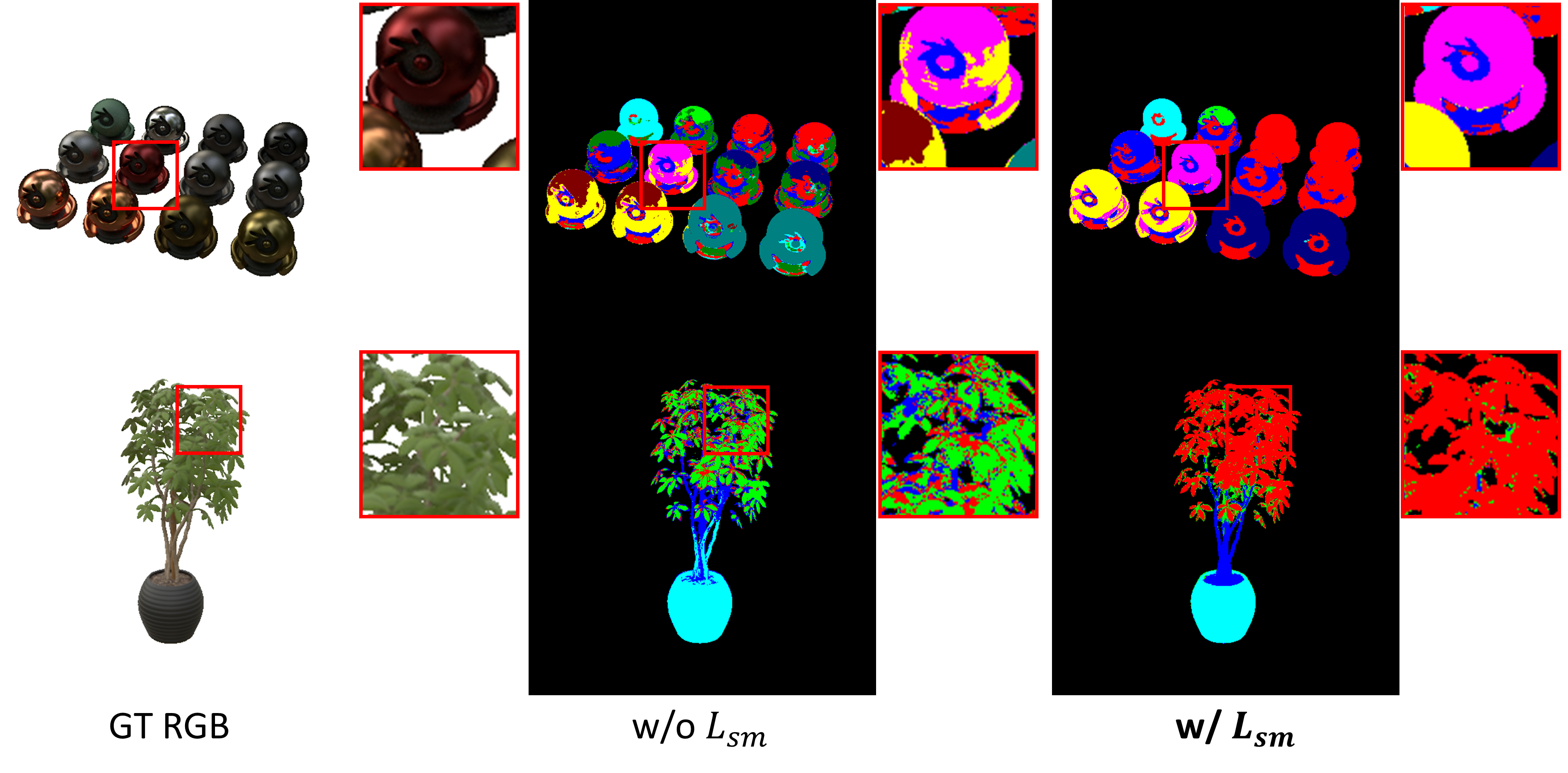}
\caption{With the help of the smooth loss, our method produces concise segmentation results, which is beneficial for convenient editing.}
\label{fig:lsm}
\end{figure}

\begin{table}[h]
\caption{Ablation on smooth loss.}
\label{tab:lsm}
\centering
\begin{tabular}{c|cccc}
\hline
\multicolumn{1}{c|}{} & \multicolumn{4}{c}{Material Clustering} \\
\hline
 & Micro-F1↑ & Macro-F1↑ & Macro-P↑ & Macro-R↑\\
w/o $L_{chr}$ & 0.779 & 0.372 & 0.348 & 0.411 \\
Ours (w/ $L_{chr}$) & {\bf 0.821} & {\bf 0.421} & {\bf 0.405} & {\bf 0.449} \\
\hline
\end{tabular}
\end{table}

\begin{figure}[h]
\centering
\includegraphics[width=0.95\linewidth]{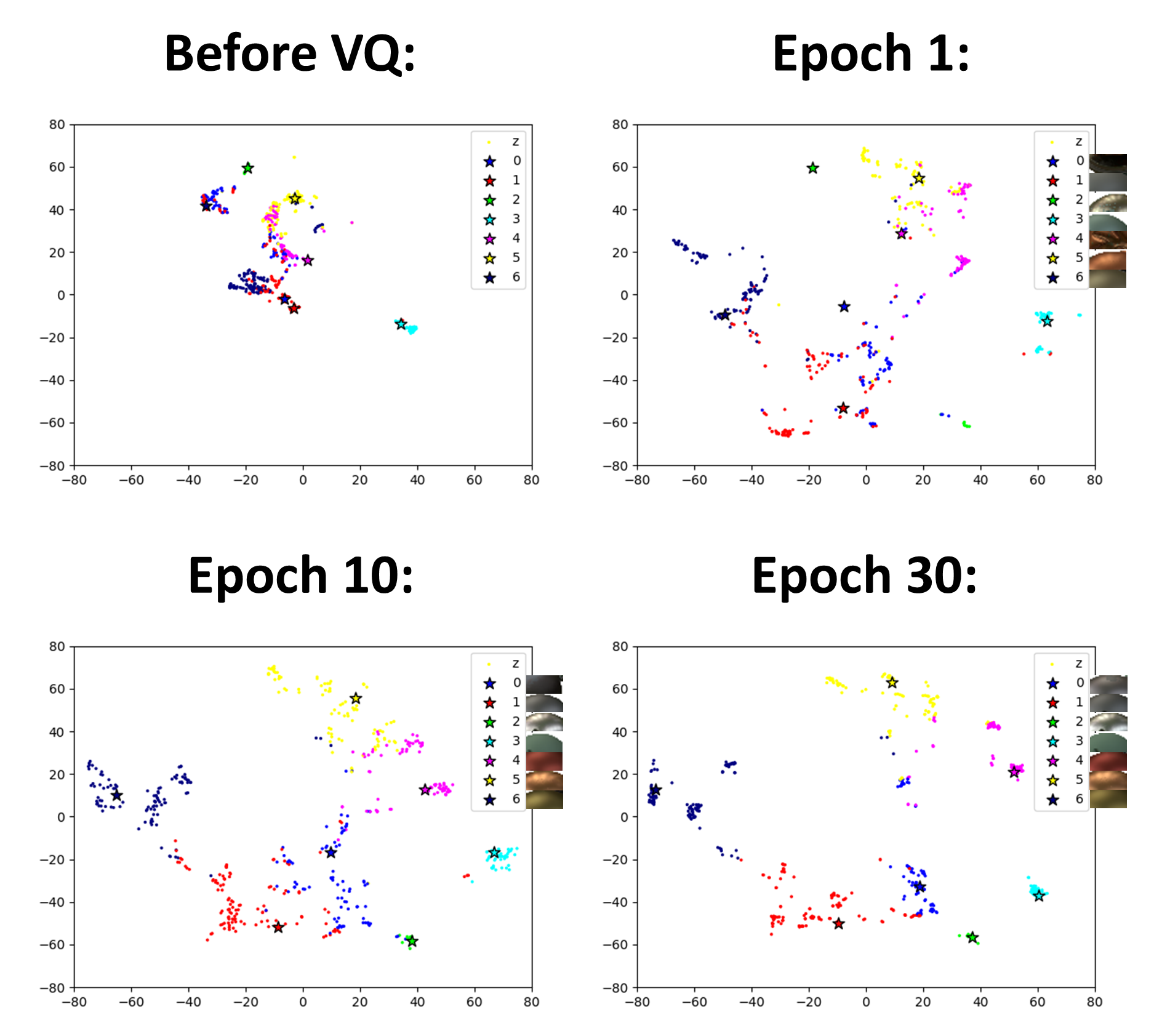}
\caption{Visualization of the model optimization using t-SNE. Dots represent continuous vectors (cluster samples), stars represent discrete VQ codewords (cluster centers). As the training progresses, different materials are gradually pulled apart and materials of the same type are compacted together. This demonstrates the effectiveness of our optimization strategy, bringing benefits such as noise suppression and segmentation enhancement.}
\label{fig14}
\end{figure}

\subsubsection{Ablation on $L_{sm}$}

To make our discrete branch produce smooth segmentation results, we introduce a weighted smooth loss $L_{sm}$. To demonstrate its effectiveness, we compare our method to a model trained without $L_{sm}$. Table \ref{tab:lsm} and Fig. \ref{fig:lsm} present the quantitative and qualitative results. Obviously, the smooth loss helps our model produce more concise segmentations, which is beneficial for convenient material editing.

\subsection{Visualization of Model Optimization}

We use t-SNE to visualize the model optimization for the \textit{metal-balls} scene. The corresponding figure is shown in Fig. \ref{fig14}. Dots represent the continuous material vectors, and stars represent the corresponding VQ codewords. Material categories are mixed before the optimization. As training progresses, individual materials are gradually pulled apart into isolated clusters. Since the latent distribution of individual materials is compacted in each cluster, the decomposition noise is suppressed accordingly. On the other hand, with the simultaneous adjustment of the material vectors (cluster samples), the VQ codewords (cluster centers) can converge to more appropriate positions, thereby improving the segmentation accuracy.